\documentclass[runningheads]{llncs}

\usepackage{eccv}

\usepackage{eccvabbrv}

\usepackage{graphicx}
\usepackage{booktabs}

\usepackage{booktabs}
\usepackage{multirow}
\usepackage{arydshln}
\usepackage{wrapfig}

\usepackage[dvipsnames]{xcolor}

\usepackage[accsupp]{axessibility}  

\usepackage{hyperref}

\usepackage{orcidlink}

\begin{document}

\title{Multi-View Foundation Models} 


\author{
    Leo Segre$^{*}$ \and  
    Or Hirschorn$^{*}$ \and Shai Avidan
}

\institute{ Tel Aviv University \\ 
\url{https://leosegre.github.io/Multi-View-Foundation-Models} 
}

\maketitle

\begin{figure}
    \centering
    \includegraphics[width=0.60\textwidth]{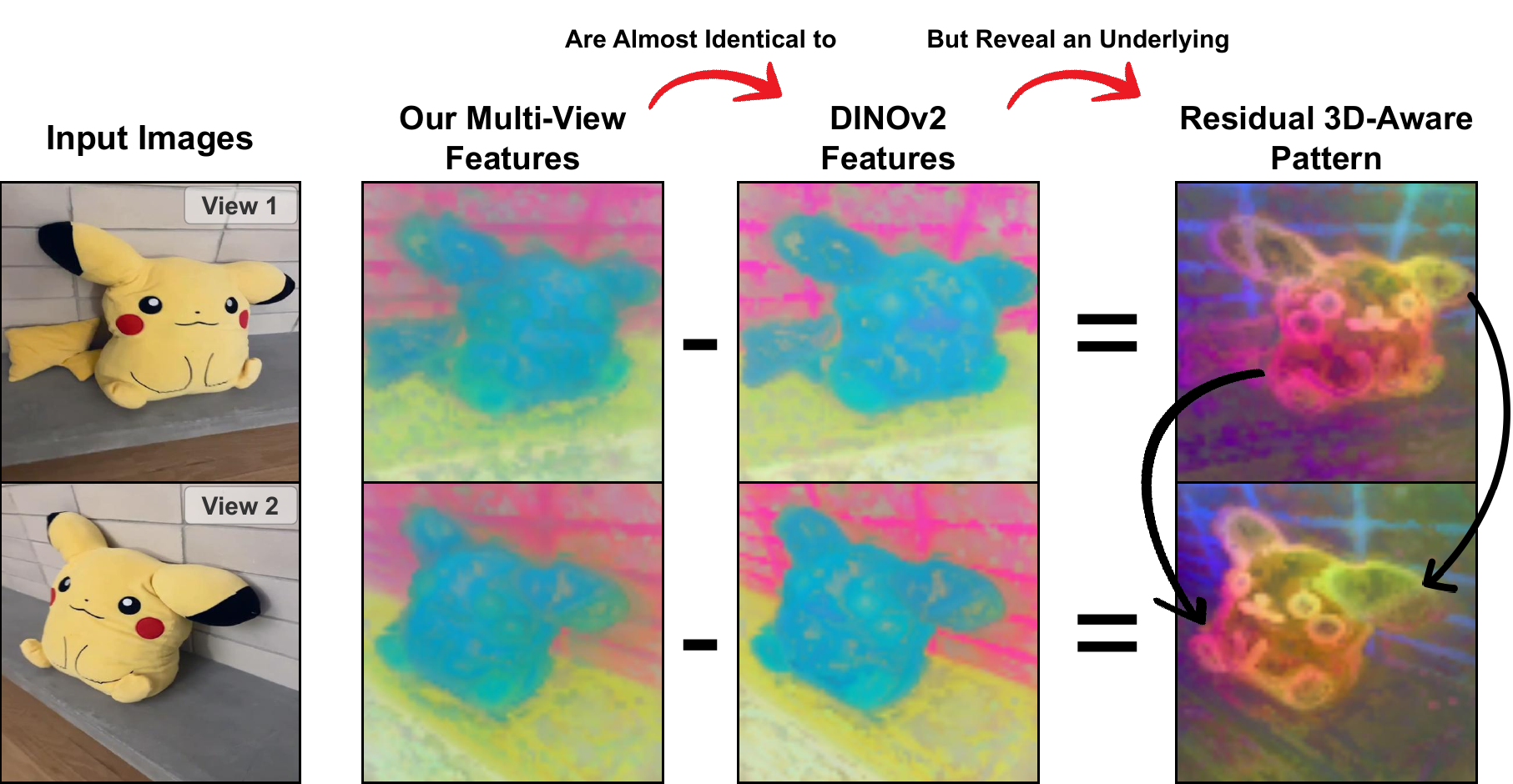}
    \hspace{0.02\textwidth}
    \includegraphics[width=0.34\textwidth]{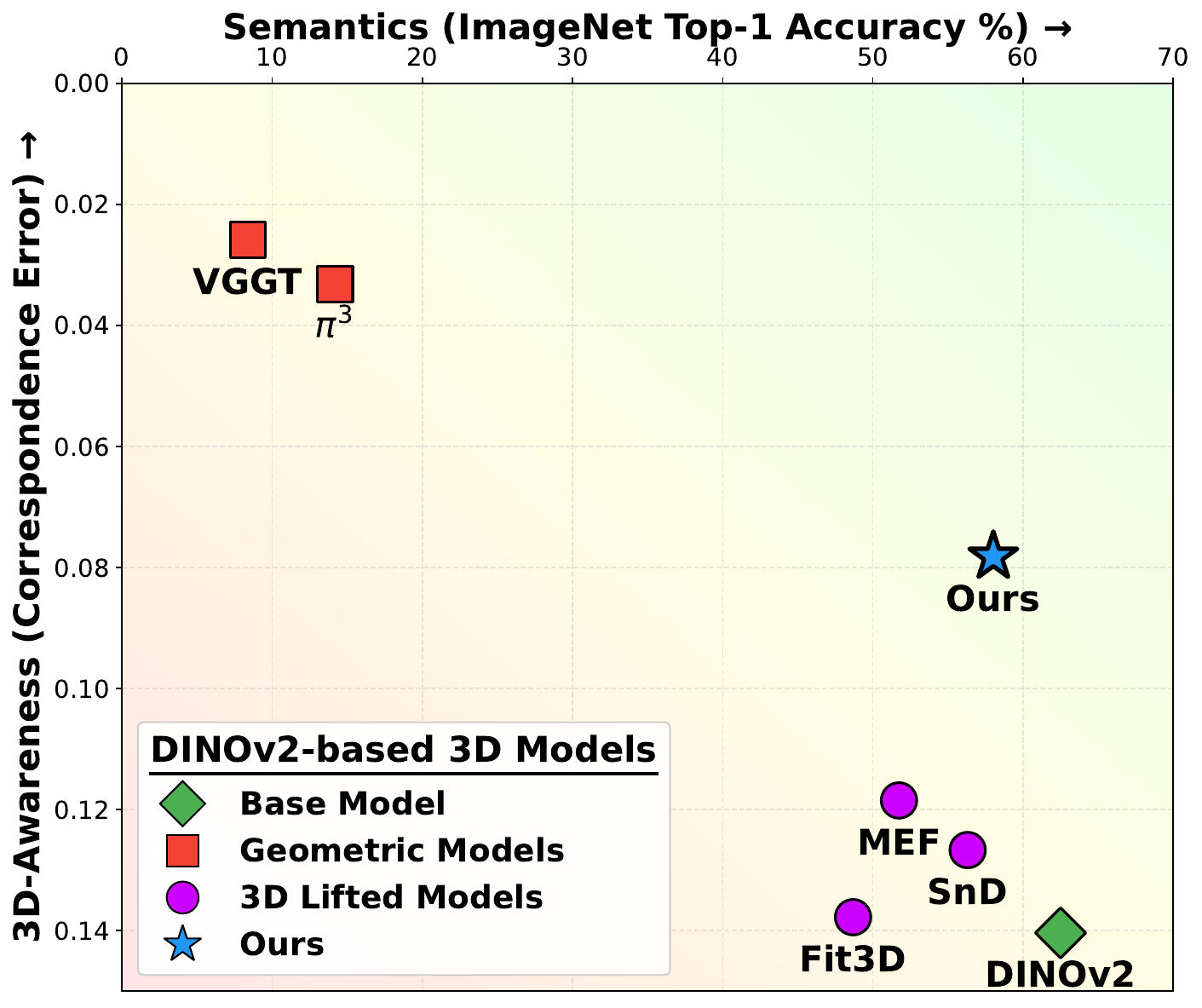}
    \caption{
    \textbf{3D-Awareness VS Semantics Trade-off.}
    (Left) Our method produces features that are semantically nearly identical to DINOv2, yet significantly more 3D-aware. Subtracting the original DINOv2 features from ours reveals a residual that distinguishes between object regions yet remains consistent across viewpoints, exposing the 3D-aware signal our method adds while preserving semantics.
    (Right) The scatter plot confirms this quantitatively: our method achieves both high semantic accuracy and low correspondence error, occupying a region of the trade-off space unreached by prior methods.
}
    \label{fig:teaser}
\end{figure}
\begingroup
\renewcommand\thefootnote{\fnsymbol{footnote}}
\footnotetext[1]{Equal contribution.}
\endgroup
\begin{abstract}
Foundation models are vital tools in various Computer Vision applications. They take as input a single RGB image and output a deep feature representation that is useful for various applications. 
However, in case we have multiple views of the same 3D scene, they operate on each image independently and do not always produce consistent features for the same 3D point. 
We propose a way to lift a Foundation Model into a Multi-View Foundation Model. Such a model takes as input a set of images and outputs a feature map for each image such that the features of corresponding points are as consistent as possible. This approach bypasses the need to build a consistent 3D model of the features.
Specifically, we show how to augment any Transformer-based foundation model (i.e., DINO, SAM, CLIP) with intermediate 3D-attention layers that help match features across different views.
Quantitative evaluations demonstrate that our approach achieves superior 3D-awareness while best preserving the native semantic space of the original foundation model, outperforming existing lifting techniques.

  \keywords{Foundation Models \and Multi-view Consistency \and 3D Lifting}

\end{abstract}    
\section{Introduction}
\label{sec:intro}

In recent years, 2D foundation models such as DINO~\cite{oquab2023dinov2}, CLIP~\cite{radford2021learning}, and SAM~\cite{kirillov2023segment} have demonstrated remarkable generalization across diverse visual tasks~\cite {li2023mask, ke2024repurposing, saxena2023surprising, amir2021deep, dino_tracker_2024}. Trained on massive datasets, they produce rich, transferable representations that excel at zero-shot and few-shot semantic understanding. Yet, despite their power, these models are "blind" to the 3D world. Because they are trained on isolated 2D images, they lack the geometric constraints necessary for spatial coherence. As a result, the same physical 3D point often yields different semantic features when viewed from different perspectives. This inconsistency creates a fundamental bottleneck for 3D-aware applications.


Our goal is to resolve this through \textbf{joint multi-view processing}, ensuring a 3D point produces the same feature across views, without sacrificing the model's native semantic intelligence. We approach this as a \textbf{feature lifting} problem: enhancing an existing foundation model with 3D-awareness while preserving its learned semantic manifold. Figure~\ref{fig:teaser} (Right) visualizes this trade-off, revealing a clear separation between lifting methods that preserve semantics (points in the bottom right corner of the figure) and end-task models that sacrifice them for geometric precision (points in the top left corner of the figure).

Early lifting approaches relied on optimization-based rendering such as NeRF or Gaussian Splatting. While consistent, these methods are slow, scene-specific, and fragile under sparse views. 
More recently, feed-forward lifting methods have been introduced to address these issues. 
For instance, FiT3D~\cite{yue2024improving2dfeaturerepresentations}, MEF~\cite{you2024multiview}, and SnD~\cite{shavin2026splat} fine-tune foundation models to achieve multi-view coherence.
However, because they still process images independently, they struggle to maintain global agreement across large viewpoint changes.
In contrast, while DINOv2-based methods like VGGT~\cite{wang2025vggt} and $\pi^3$~\cite{wang2025pi} excel as geometric experts, they recover explicit geometry at the expense of the semantic manifold. This leads to "catastrophic forgetting" of the zero-shot capabilities that make foundation models valuable.

In this paper, we introduce a method that shifts the "lifting" paradigm by processing multiple images jointly. We "inflate" the foundation model to natively handle multi-view inputs in a single forward pass, enabling the model to see all viewpoints at once.
The key challenge in this approach is adding multi-view awareness without distorting the model's original semantic space - a balance that previous methods fail to maintain. 
We address this by interleaving multi-view adapters directly within the pre-trained layers, thereby facilitating seamless cross-view communication while preserving the original feature manifold. Thus, our approach (middle right point in the scatter plot in Figure~\ref{fig:teaser}) offers a better trade-off of consistency and semantics than previous "lifting" methods or geometry reconstruction methods. Figure~\ref{fig:teaser} (Left) visualizes the features generated by our method. They are close to the original DINOv2 features, thus maintaining their semantic power, while introducing residual features that remain consistent across viewpoints. 

Furthermore, our architecture is paired with a scalable training pipeline designed for direct feature-space adaptation. 
Existing methods often enforce 3D consistency through computationally expensive bottlenecks. Approaches relying on explicit 3D modeling require either heavy per-scene optimization or complex feed-forward 3D reconstruction to render feature targets. 
Conversely, MEF~\cite{you2024multiview} utilizes multi-view correspondences, but still relies on indirect, highly engineered loss formulations to constrain the features. 
We bypass these complications entirely. Our training relies strictly on readily available multi-view correspondences (e.g., extracted via COLMAP) to directly guide cross-view feature alignment. By decoupling our geometric priors from explicit 3D modeling and complex training objectives, our pipeline eliminates massive per-scene, per-backbone overheads, allowing us to efficiently adapt any foundation model.

We evaluate our method across diverse benchmarks and challenging datasets such as ScanNet++~\cite{yeshwanth2023scannet}, focusing on two complementary criteria: gaining 3D awareness and maintaining original semantics. 
First, we assess 3D awareness, as defined by Banani~\etal~\cite{banani2024probing}, to quantify the spatial coherence missing in 2D models. Following their evaluation protocol, we demonstrate significant improvements in feature correspondence and surface normal estimation. 
Second, we confirm that our framework preserves the semantic manifold by measuring feature-space cosine similarity relative to the base models and achieving on-par performance in zero-shot classification. 
These results generalize across real and synthetic scenes and multiple backbones, including DINOv2, DINOv3, CLIP, and SAM, confirming that joint multi-view processing provides the spatial coherence that single-view models inherently lack.


Our contributions are summarized as follows:
\begin{itemize}
    \item We introduce a novel feature lifting framework that "inflates" 2D foundation models to handle multiple views in a single forward pass, striking an effective balance between spatial coherence and semantic preservation.
    \item We propose a scalable training pipeline that eliminates the current overheads of explicit 3D modeling and indirect loss constraints. By utilizing readily available geometric correspondences, we enforce multi-view consistency directly in feature space.
    \item We demonstrate that our scalable training scheme generalizes seamlessly across diverse foundation models (DINOv2, DINOv3, CLIP, SAM) and scene types, achieving state-of-the-art results on different benchmarks.
\end{itemize}

\section{Related Works}
\label{sec:related}

\subsection{3D-Aware Foundation Models}
The unprecedented success of 2D foundation models has inspired a concerted effort to extend their semantic knowledge into the 3D domain. A fundamental obstacle, however, is that their learned representations are inherently viewpoint-dependent and lack an explicit understanding of 3D geometry. This limitation has been empirically confirmed in recent studies. For instance, Banani~\etal~\cite{banani2024probing} conducted a thorough analysis demonstrating that features from prominent vision models, including DINO~\cite{oquab2023dinov2}, CLIP~\cite{radford2021learning}, SAM~\cite{kirillov2023segment}, and Stable Diffusion~\cite{rombach2022high}, fail to remain consistent across different views of the same object.
Recent research has therefore focused on explicitly lifting 2D features into coherent, 3D-aware representations. A prevalent approach is per-scene test-time optimization, in which a 3D scene representation, such as a Neural Radiance Field (NeRF)~\cite{mildenhall2020nerf}, 3D Gaussian Splatting (3DGS)~\cite{kerbl20233d}, or voxel-based feature grid~\cite{li2023dfa3d, sun2022direct, yu2021plenoxels}, is optimized to align with 2D features \cite{Ye_2023, Zhi:etal:ICCV2021, siddiqui2022panopticlifting3dscene, qiu-2024-featuresplatting, zhou2024feature3dgssupercharging3d}. This is typically achieved by minimizing a feature-matching loss between renderings from the 3D model and features extracted from multiple input views. While these methods yield geometrically consistent feature fields, their reliance on costly, per-scene optimization makes them impractical for large-scale applications.

To overcome this efficiency bottleneck, another promising approach is to focus on building inference-based methods that replace the optimization process with a single feed-forward network.
Lift3D~\cite{t2024lift3d} tackled the issue of per-scene (or per-task) optimization by using generalizable novel-view synthesis: they treat the 2D feature maps as colors and render feature maps of new views via a NeRF-inspired pipeline, integrating multi-view geometry and RGB cues to enforce consistency. However, their method assumes an encoder–decoder structure for the 2D foundation model, and still requires rendering novel views, which is computationally heavy.
Later, FiT3D~\cite{yue2024improving2dfeaturerepresentations} made a major step forward by fine-tuning a 2D foundation model to directly predict a canonical 3D feature field from a single input image. This makes processing much faster by replacing repeated optimization with a single inference step. 
Snd~\cite{shavin2026splat} expanded this approach, using feed-forward 3DGS to bypass per-scene optimization by directly attaching lifted 2D features to Gaussian primitives. This allows for the efficient rendering of 3D-consistent feature maps at novel views, providing a dense geometric signal that supervises a student model through a scalable, EMA-driven distillation paradigm.
Alternatively, MEF~\cite{you2024multiview} uses Smooth-AP~\cite{brown2020smooth} loss to encourage multi-view correspondence. It employs LORA fine-tuning and adds a single convolution layer to the original model architecture. Training is done on synthetic 3D assets, which provide corresponding pixels.
Yet, all these methods process images individually at inference time - limiting their ability to maintain multi-view consistency. Our work addresses this limitation by introducing a framework that processes multiple views together to create a unified and multi-view consistent representation.

\subsection{Inflating 2D Models}
A dominant paradigm in 3D~\cite{karnewar2023holodiffusion, chan2023generative, poole2022dreamfusion, lin2023magic3d} and video~\cite{blattmann2023stable, yang2024cogvideox, ho2022video} generation is “inflating” powerful, pre-trained 2D vision models. Most state-of-the-art diffusion frameworks are built upon 2D image diffusion backbones, and share a common principle: they inflate the 2D layers (convolutions or attention) into 3D, modifying the architecture while preserving the pre-trained 2D knowledge acquired from large-scale image datasets. 
These inflated models are then fine-tuned to produce outputs that are temporally and spatially consistent in 3D.

Beyond generative tasks, this inflation principle has recently been applied to representation learning. Kim~\etal~\cite{kim2025exploring} introduced temporal attention layers into DINOv2 and fine-tuned the model for point tracking, yielding temporally consistent DINOv2 features. 
Together, these works highlight a rapidly growing trend: developing general mechanisms that imbue powerful 2D models with awareness of higher-order dimensions such as time and 3D geometry. Our work is situated within this trend, focusing specifically on inflating 2D foundation models to create multi-view consistent representations, while preserving the semantic manifold as much as possible.

\section{Method}
\begin{figure*}[t]
    \centering
    \includegraphics[width=0.98\linewidth]{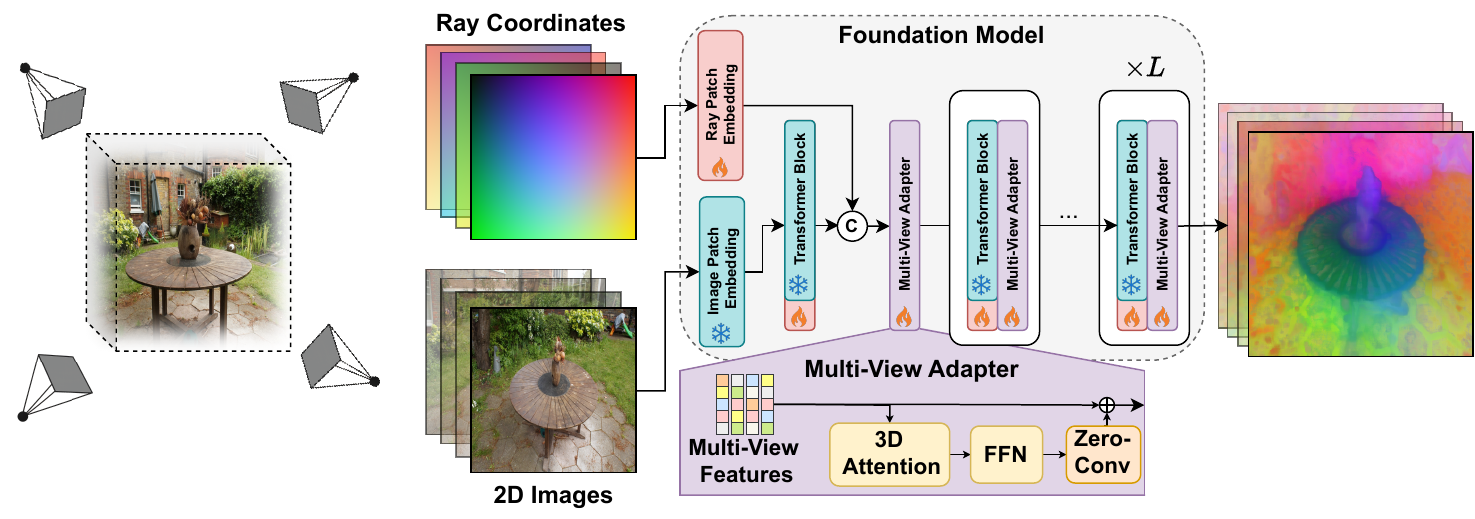}
    \caption{\textbf{Framework Overview.} 
     Our method integrates a pre-trained foundation model with multi-view spatial adapters denoted in \textcolor{purple}{purple} for 3D-aware feature learning. These adapters are added after each Transformer block.
     Given $M$ input images with camera poses, per-view features are extracted and fused via 3D-aware adapter blocks conditioned on ray-based pose embeddings, producing geometry-consistent representations across views. 
    }
    \label{fig:arch}
\end{figure*}

Given a set of images of a static scene and their corresponding camera parameters, our goal is to extract 3D-aware features that preserve the semantic representations of the pretrained foundation model. 
We begin by describing the design of our multi-view adapter and its integration with a pretrained transformer backbone. We then present our training strategy for learning 3D-aware features across multiple views.




\subsection{Multi-View Adapter}

The primary challenge in lifting 2D foundation models is introducing multi-view consistency without destroying the rich, carefully learned semantic priors of the pretrained model. 
To address this, we propose a minimally invasive adapter architecture, $A_{\phi}$, inserted after each transformer block. 
Notably, rather than applying alternating-attention as a separate module as in VGGT~\cite{wang2025vggt}, our approach interleaves global attention directly within the pre-trained layers.
This design relies on two key principles to preserve the original semantic space: alternating attention to decouple intra-view spatial learning from inter-view geometric reasoning, and zero-convolution merging to guarantee an identity mapping at initialization. Our framework is illustrated in Figure~\ref{fig:arch}. 

Formally, let $M$ denote the number of input views. 
At layer $l$, the input to the block is a tensor of latent feature maps $\mathbf{Z}^{l-1} \in \mathbb{R}^{M \times (H \times W) \times C}$, where $H \times W$ represents the flattened spatial dimensions and $C$ is the channel dimension. 
First, the base 2D transformer block, $f_{\theta}^l$, processes each of the $M$ views independently. This block applies standard spatial self-attention to extract intra-view semantics, outputting intermediate 2D features $\tilde{\mathbf{Z}}^l$:
\begin{equation}
\tilde{\mathbf{Z}}^l_i = f_{\theta}^l(\mathbf{Z}^{l-1}_i) \quad \forall i \in \{1, \dots, M\}
\end{equation}

To support multi-view reasoning, we "inflate" the 2D foundation model by interleaving 3D-attention adapters $A_{\phi}^l$. These modules enable global communication by treating all patches across all $M$ views as a single joint sequence, allowing every patch to attend to every other patch in the multi-view set. Formally, we flatten $\tilde{\mathbf{Z}}^l_i \in \mathbb{R}^{(M \times H \times W) \times C}$ and use standard transformer block to process multi-view information:
\begin{equation}
\mathbf{H}^l_{1:M} = A_{\phi}^l({\mathbf{Z}}^l_{1:M}) = \text{FFN}(\text{MSA}(\tilde{\mathbf{Z}}^l_{1:M}))
\end{equation}
where MSA denotes Multi-head Self-Attention. To preserve the pretrained feature manifold at initialization, the result is integrated back into the main stream via a zero-initialized linear projection:
\begin{equation}
\mathbf{Z}^l_{i} = \tilde{\mathbf{Z}}^l_{i} + \text{ZeroProj}(\mathbf{H}^l_{i})
\end{equation}
This "zero-gate" ensures that at the start of training, the inflated model’s behavior is identical to its 2D backbone.

To provide the necessary geometric context for multi-view reasoning, we construct explicit ray embeddings. For each view $i$, we compute a dense raymap based on its camera pose relative to the first view. The Plücker coordinates for a spatial token at pixel location $(u,v)$ are defined as:
\begin{equation}
\mathbf{p}_i(u,v) = [\mathbf{o}_i(u,v), \mathbf{d}_i(u,v)] \in \mathbb{R}^6
\end{equation}
where $\mathbf{o}$ and $\mathbf{d}$ denote the ray origin and direction. These raymaps are projected into the latent channel dimension $C'$ via a patch embedding network to form the geometric conditioning tensor $\mathbf{P} \in \mathbb{R}^{M \times (H \times W) \times C'}$.
These are then concatenated with the initial intermediate features $\tilde{\mathbf{Z}}^0$ channel-wise.

During training, we employ LoRA~\cite{hu2022lora} fine-tuning for the original foundation model weights ($\theta$), alongside full fine-tuning of the adapter parameters ($\phi$). This scalable training scheme yields superior geometric performance while better preserving the generalization capacity of the pre-trained foundation model.

\subsection{3D-Aware Training}

Let $F_i$ and $F_j$ denote the output feature maps (e.g., the final layer outputs $\mathbf{Z}^L_i$ and $\mathbf{Z}^L_j$) for view $i$ and view $j$, respectively.
We denote corresponding spatial locations as $x_{i,k}$ in view $i$ and $x_{j,k}$ in view $j$, where $k$ indexes a specific matched point across the views.
Naively enforcing feature consistency via a simple L2 loss, e.g.,
\begin{equation}
\mathcal{L}_{\text{naive}} = \| F_i(x_{i,k}) - F_j(x_{j,k}) \|_2
\end{equation}
is problematic because the trivial solution $F_i(x) = F_j(x) = C$ globally minimizes the objective, leading to feature collapse. This reveals that multi-view consistency alone is insufficient - the model 
must learn to produce features that are not only consistent across views, but 
also spatially discriminative within each view. In other words, the features 
must be \textbf{3D-aware}.

To avoid this, and encourage consistent geometric reasoning across views, we adopt the geometry-aware dense loss proposed by Zhang~\etal~\cite{zhang2024telling}. 
This loss was introduced to learn geometry-aware feature representations that remain sensitive to 3D orientation and spatial relationships, addressing the limitation of traditional sparse semantic correspondence losses.

We start by computing a similarity map between the normalized query feature $F_i(x_{i,k})$ and all spatial locations in the target feature map $F_j$:
\begin{equation}
S^{i\rightarrow j}_k(u) = \frac{F_i(x_{i,k})}{\|F_i(x_{i,k})\|} \cdot \frac{F_j(u)}{\|F_j(u)\|}, \quad \forall u \in \Omega_j
\end{equation}
where $\Omega_j$ denotes the spatial domain of $F_j$. 
We then convert $S^{i\rightarrow j}_l$ into a probability distribution using a softmax with temperature $\tau$ and apply a differentiable SoftArgMax to obtain the predicted correspondence location:
\begin{equation}
P^{i\rightarrow j}_k(u) = \mathrm{softmax}\!\left(\frac{S^{i\rightarrow j}_k(u)}{\tau}\right), \quad \hat{x}_{j,k} = \sum_{u \in \Omega_j} P^{i\rightarrow j}_k(u) \, u
\end{equation}
Finally, the geometry-aware dense loss penalizes the Euclidean distance between the predicted and ground-truth correspondence:
\begin{equation}
\mathcal{L}_{\text{corr}} = \| x_{j,k} - \hat{x}_{j,k} \|_2
\end{equation}

This correspondence loss is computed over multiple corresponding points and multiple view pairs in each training batch, using correspondences derived from 3D point clouds. 
By enforcing geometric consistency in the feature space, this objective prevents trivial collapse while effectively bridging the gap between 2D pretrained representations and multi-view 3D-aware features.

We additionally incorporate a regularization term to prevent the refined embeddings from deviating excessively from the original feature space of the pretrained model. 
Following~\cite{tumanyan2024dino}, we encourage each refined feature to (1) maintain a high cosine similarity and (2) preserve a similar norm to its corresponding original feature. This regularization helps retain the strong generalization ability of the foundation model:

\begin{equation}
\mathcal{L}_{\text{reg}} = 
\underbrace{1 - \frac{\|\tilde{F}_i\|}{\|F_i\|}}_{\mathcal{L}_{\text{norm}}} + 
\underbrace{1 - \cos\!\left( \tilde{F}_i, F_i \right)}_{\mathcal{L}_{\text{angle}}}
\end{equation}
where $\tilde{F}_i$ and $F_i$ denote the original (pretrained) and adapted features for view $i$, respectively.
The total training objective combines both terms:
\begin{equation}
\mathcal{L}_{\text{total}} = \mathcal{L}_{\text{corr}} + \lambda_{\text{reg}} \, \mathcal{L}_{\text{reg}}
\end{equation}
with $\lambda_{\text{reg}}$ controlling the strength of the regularization.

A key advantage of this training formulation is its scalability. Unlike previous approaches that require per-scene optimization via 3D representations like Gaussian Splatting (GS) or Neural Radiance Fields (NeRF) for every new foundation model, our method decouples the geometry from the feature space.
Because $\mathcal{L}_{\text{corr}}$ is defined directly over spatial coordinates derived from a fixed correspondence dataset, the same supervision signal can be seamlessly applied to adapt any 2D foundation model. This eliminates the need for expensive per-scene re-optimization or model-specific geometric reconstruction, allowing for efficient fine-tuning across diverse model architectures (e.g., DINOv2, SAM, or CLIP) using a unified training pipeline and dataset.

\section{Experiments}
We evaluate our approach along two complementary axes:

\noindent\textbf{Semantic preservation.} We measure how faithfully the original foundation model capabilities are retained. Specifically, we report (i)~cosine similarity between our features and the original foundation model features, and (ii)~zero-shot KNN classification accuracy - a purely semantic, single-view benchmark that is agnostic to any 3D reasoning.

\noindent\textbf{3D-awareness.} We measure how well the learned features capture geometric structure across views. We evaluate on (i)~a correspondence task, where ground-truth matches are obtained by projecting 3D point clouds onto 2D frames, and (ii)~3D-aligned surface normal estimation. We note that these evaluations are designed to measure intrinsic feature quality rather than to compete with task-specific state-of-the-art methods. As we only use simple feature probing, these tasks serve as indicators of how promising the backbone is as a 3D-aware foundation model.

\noindent\textbf{Model Variants.} We evaluate two primary versions of our framework: a full pipeline utilizing camera poses and a pose-free variant that omits ray embedding concatenation. This provides a trade-off between performance and applicability in scenarios where camera data is unavailable. 
We also demonstrate that our framework readily generalizes to additional transformer-based foundation models beyond DINOv2, without any architectural modification.

\paragraph{\textbf{Datasets.}}
We train on ScanNet++ (230 scenes, 140K views). For correspondence evaluation, we test on 50 held-out ScanNet++ scenes and a generalization set of 115 scenes from eight diverse datasets: MipNeRF-360~\cite{barron2022mipnerf360}, LLFF~\cite{mildenhall2019llff}, Tanks \& Temples~\cite{knapitsch2017tanks}, NeRF-Synthetic~\cite{mildenhall2020nerfrepresentingscenesneural}, ScanNet~\cite{dai2017scannet}, Objaverse~\cite{deitke2022objaverse}, Nerfbusters~\cite{Nerfbusters2023}, and NeRF2NeRF~\cite{10160794}, spanning indoor, outdoor, real, synthetic, and object-centric scenarios. 
For surface normal prediction, we use NAVI~\cite{jampani2023navi} (34 high-quality meshes with multi-view images), rendering ground-truth 3D-aligned normals following~\cite{banani2024probing}.

\paragraph{\textbf{Baselines.}}
We compare against feed-forward "lifting" methods that enhance features with geometric awareness: MEF~\cite{you2024multiview}, FiT3D~\cite{darcet2024visiontransformersneedregisters}, and SnD~\cite{shavin2026splat}. All operate on the small DINOv2 backbone, isolating the effect of each lifting strategy.

\paragraph{\textbf{Implementation Details.}}
\textbf{Training -} We train on ScanNet++ for 24 epochs (12 for SAM and DINOv3), sampling 1{,}000 sets of 4 images (512$\times$512) per scene per epoch. At each iteration, we compute the correspondence loss over 128 sampled correspondences. We use AdamW ($lr=10^{-4}$) with a linear scheduler and apply LoRA fine-tuning with rank $r{=}32$ and $\alpha{=}32$.
\textbf{Backbones -} We evaluate on DINOv2 ViT-S/14 with register tokens~\cite{darcet2024visiontransformersneedregisters}, SAM ViT-B, DINOv3 ViT-B/16, and CLIP ViT-B/16.
\textbf{Probing -} To assess 3D consistency, we freeze all backbone weights and train a lightweight self-attention-based head to predict per-pixel normals in a shared 3D reference frame (camera 0 as origin). Following Banani~\etal\cite{banani2024probing}, the probe outputs four channels $(x,y,z,\sigma)$ trained with an uncertainty-weighted loss. Training uses AdamW for 50 epochs ($lr=5{\times}10^{-5}$).

\begin{table}[t]
\centering
\small
\setlength{\tabcolsep}{4pt}
\caption{
\textbf{Semantic Preservation.}
Cosine similarity to the original DINOv2~\cite{oquab2023dinov2} features and zero-shot KNN classification accuracy on ImageNet~\cite{deng2009imagenet}.
}
\begin{tabular}{lcccc}
\toprule
& \multicolumn{2}{c}{\textbf{DINOv2 Similarity}~($\uparrow$)} & \multicolumn{2}{c}{\textbf{Zero-Shot Classification}~($\uparrow$)} \\
\cmidrule(lr){2-3} \cmidrule(lr){4-5}
& ScanNet++ & Generalization & Top-1 & Top-5 \\
\midrule
DINOv2~\cite{oquab2023dinov2}
& --      & --
& 62.50\% & 84.37\% \\
\hdashline
MEF~\cite{you2024multiview}
& -0.0249  & 0.0071
& 51.74\% & 74.44\% \\
FiT3D~\cite{darcet2024visiontransformersneedregisters}
& 0.7484  & 0.6111
& 48.68\% & 71.74\% \\
SnD~\cite{shavin2026splat}
& 0.6125  & 0.6390
& 56.31\% & 80.37\% \\
\midrule
\textbf{Ours - Pose-Free}
& 0.9360  & \textbf{0.8099}
& \textbf{58.02\%}     & 80.91\% \\
\textbf{Ours}
& \textbf{0.9376} & 0.8003
& \textbf{58.02\%} & \textbf{81.21\%} \\
\bottomrule
\end{tabular}
\label{tab:semantic_preservation}
\end{table}

\subsection{Semantic Preservation}

\begin{figure}[t]
    \centering
    \small
    \renewcommand{\arraystretch}{1.5} 
    \setlength{\tabcolsep}{2pt} 
    
    \begin{tabular}{ccccc}
        \textbf{Image} & \textbf{Ours} & \textbf{DINOv2} & \textbf{FiT3D} & \textbf{MEF} \\
        
        \includegraphics[width=0.18\textwidth]{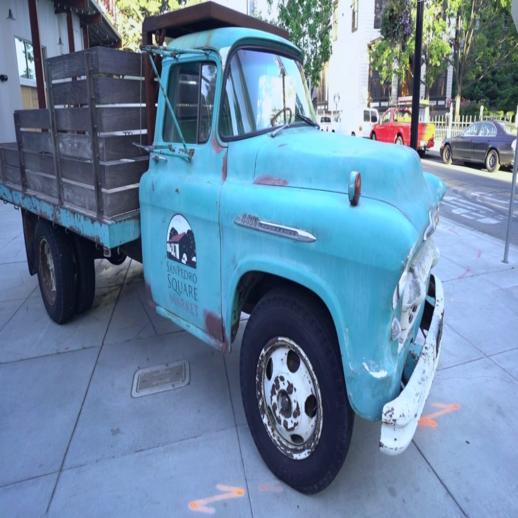} &
        \includegraphics[width=0.18\textwidth]{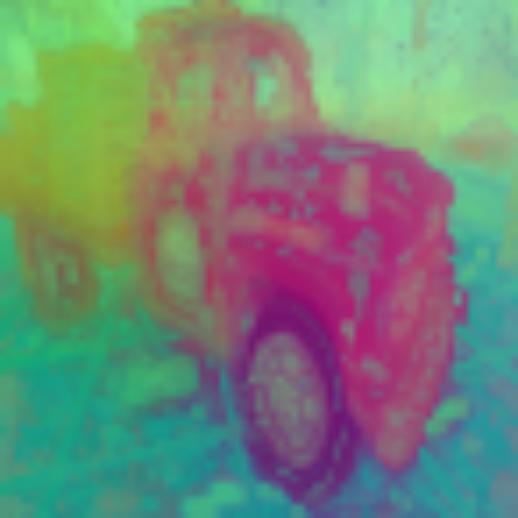} &
        \includegraphics[width=0.18\textwidth]{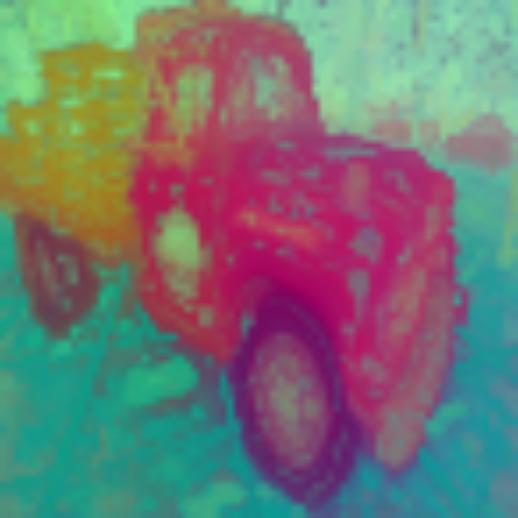} &
        \includegraphics[width=0.18\textwidth]{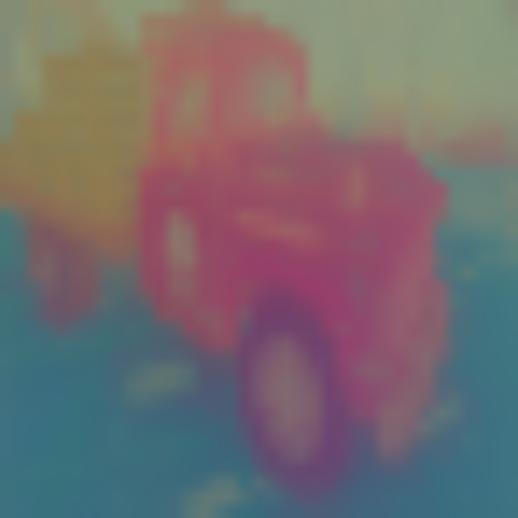} &
        \includegraphics[width=0.18\textwidth]{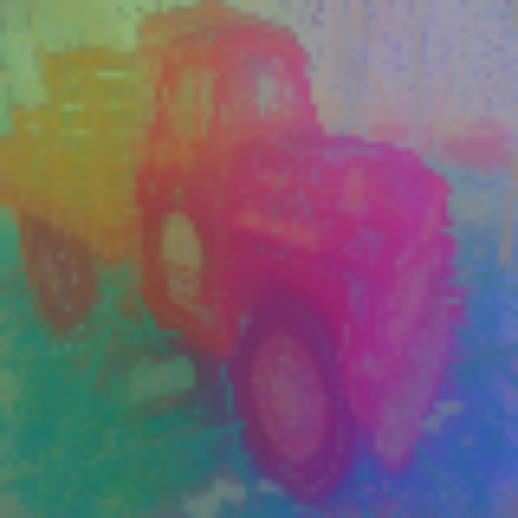} \\
        
        \includegraphics[width=0.18\textwidth]{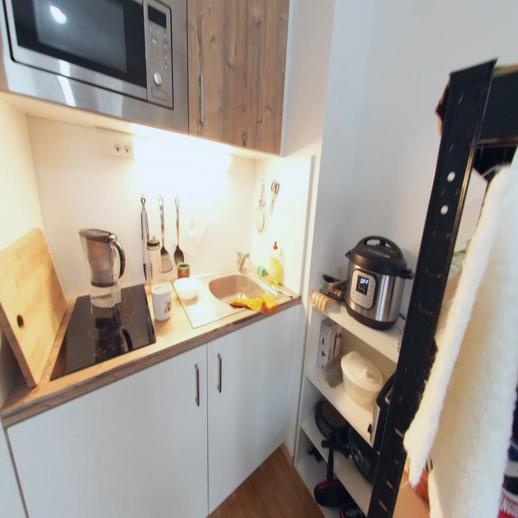} &
        \includegraphics[width=0.18\textwidth]{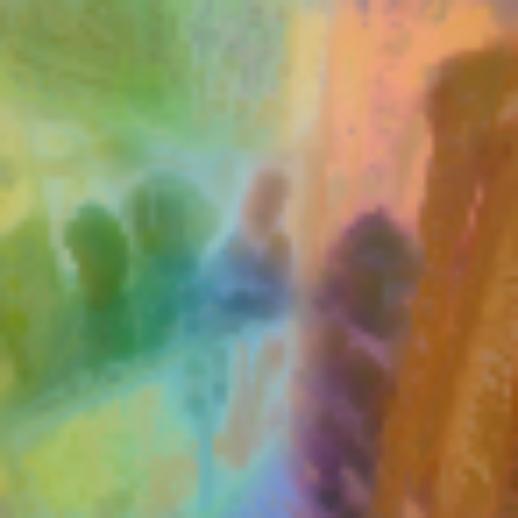} &
        \includegraphics[width=0.18\textwidth]{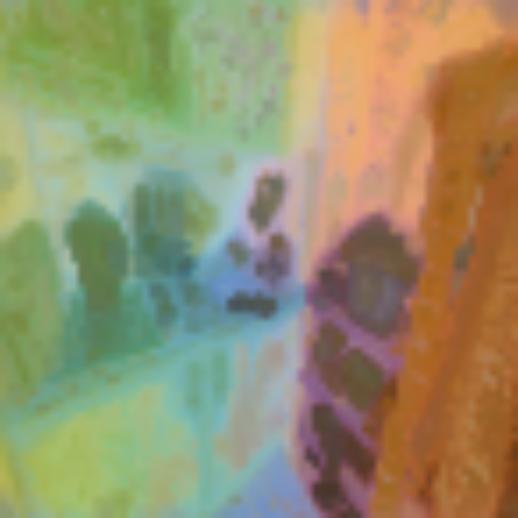} &
        \includegraphics[width=0.18\textwidth]{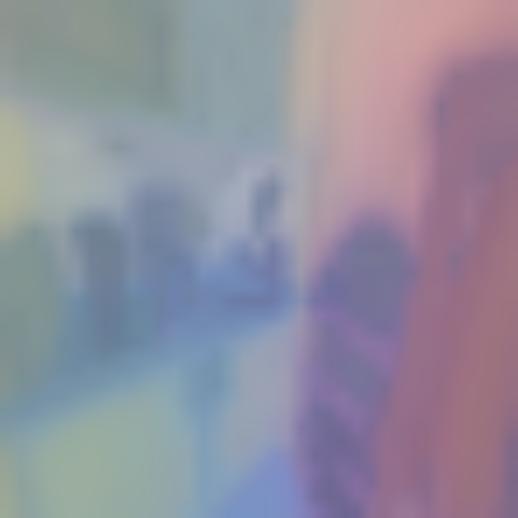} &
        \includegraphics[width=0.18\textwidth]{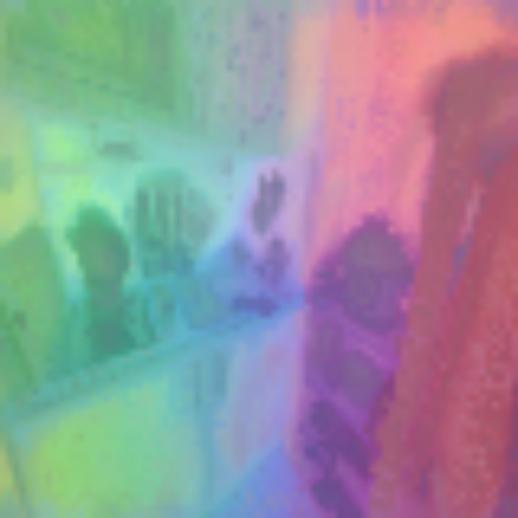} \\
    \end{tabular}
    \caption{
    \textbf{Feature Similarity to Base Model.} 
    For each input image (left), we visualize the feature embeddings of Ours, DINOv2, FiT3D, and MEF after projecting all features into a \textbf{shared PCA space}.
    The strong visual alignment between our method and the original DINOv2 shows that our multi-view adaptation preserves the semantic structure of the pretrained model, while other methods drift further away. This confirms that our method achieves 3D consistency without sacrificing fidelity to the base features. 
    }
    \label{fig:similarity_to_base}
\end{figure}

Quantitative results are provided in Table~\ref{tab:semantic_preservation}. Our method achieves the highest cosine-similarity to the original DINOv2 features across both benchmarks, substantially outperforming all baselines.
This is further reflected in zero-shot classification, where our method retains accuracy close to the original DINOv2, while other methods show a more noticeable drop. Notably, FiT3D achieves higher feature similarity than MEF yet performs worse on classification, suggesting that cosine similarity alone does not fully capture semantic capabilities and that classification provides a complementary view. 
Together, these metrics show that our model retains most of its native representational capacity after adaptation.
Figure~\ref{fig:similarity_to_base} visualizes this further: our adapted features remain visually indistinguishable from the original DINOv2 features, confirming that the pretrained manifold is largely preserved.

\subsection{3D-Awareness}

\paragraph{\textbf{Correspondence.}} For correspondence evaluation, we project 3D points reconstructed by COLMAP~\cite{schoenberger2016sfm} onto pairs of views to obtain matches. Given a point in one image, we retrieve its correspondence by finding the nearest feature in the other view. The localization error is the normalized Euclidean distance between the predicted and ground-truth locations. 

\begin{figure}[t]
\begin{minipage}{0.48\linewidth}
\centering
\small
\setlength{\tabcolsep}{4pt}
\captionof{table}{
\textbf{Correspondence Localization.}
Localization error on ScanNet++~\cite{yeshwanth2023scannet} and a generalization set spanning diverse datasets. Lower is better.
}
\resizebox{\textwidth}{!}{%
\begin{tabular}{lcc}
\toprule
& \multicolumn{2}{c}{\textbf{Loc.\ Err.}~($\downarrow$)} \\
\cmidrule(lr){2-3}
& ScanNet++ & Generalization \\
\midrule
DINOv2~\cite{oquab2023dinov2}
& 0.1029  & 0.1404 \\
\hdashline
MEF~\cite{you2024multiview}
& 0.0978  & 0.1232 \\
FiT3D~\cite{darcet2024visiontransformersneedregisters}
& 0.0858  & 0.1378 \\
SnD~\cite{shavin2026splat}
& 0.0876  & 0.1267 \\
\midrule
\textbf{Ours - Pose-Free}
& 0.0262  & 0.0861 \\
\textbf{Ours}
& \textbf{0.0247} & \textbf{0.0782} \\
\bottomrule
\end{tabular}
}
\label{tab:correspondence}
\end{minipage}
\hfill
\begin{minipage}{0.48\linewidth}
\centering
\includegraphics[width=0.98\linewidth]{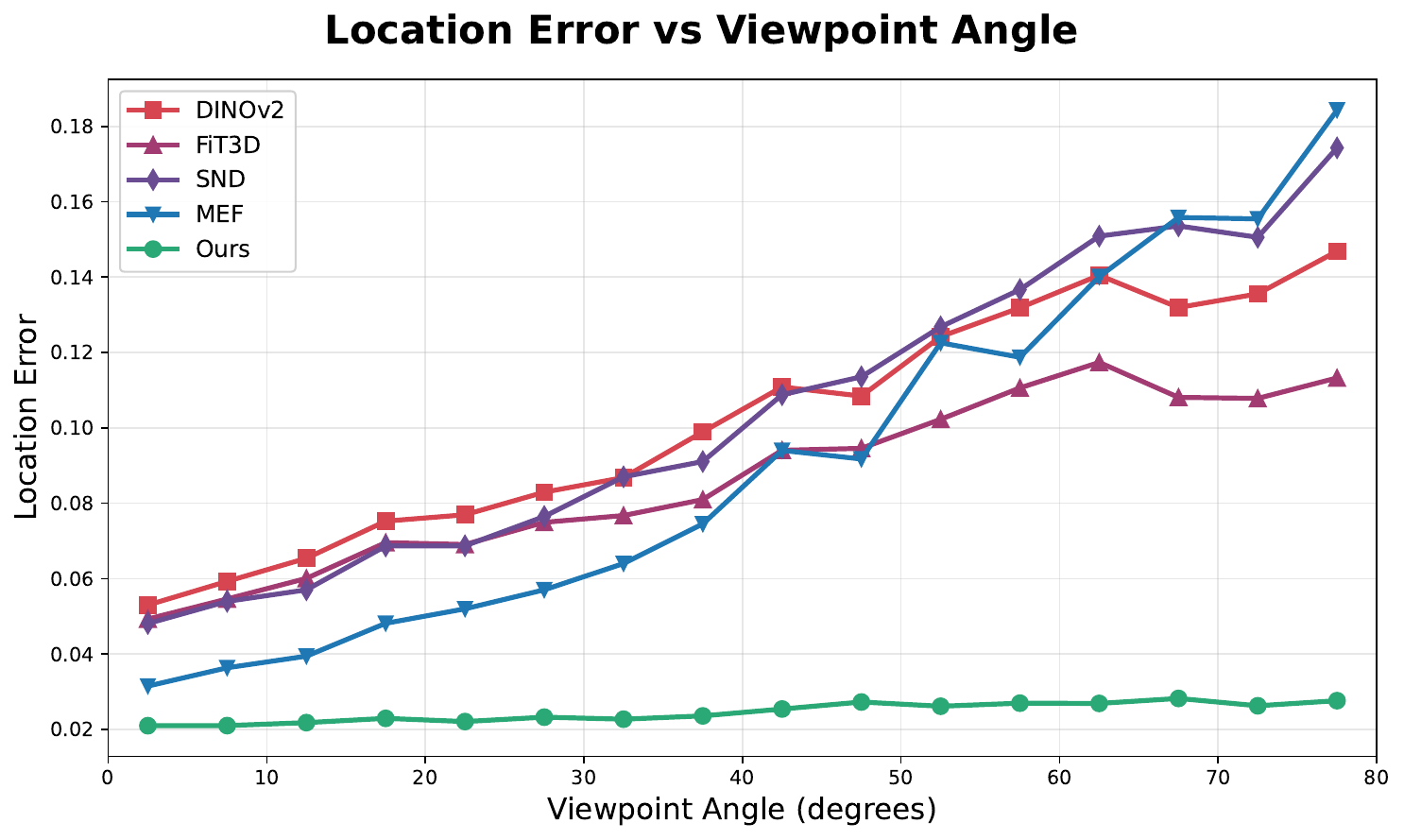}    
\caption{
\textbf{Viewpoint Robustness.} Our method maintains low error
under large viewpoint changes, outperforming other 3D-lifting methods.    
}
\label{fig:viewpoint}
\end{minipage}
\end{figure}

As shown in Table~\ref{tab:correspondence}, our method achieves a substantial improvement in correspondence localization over all baselines on both ScanNet++ and the generalization set. 
While existing methods offer only marginal gains over the original DINOv2 features, our approach reduces the localization error by a large margin. Notably, FiT3D improves on ScanNet++ but fails to generalize, whereas our method maintains strong performance across diverse unseen scenes.
This robustness is further confirmed by the viewpoint analysis in Figure~\ref{fig:viewpoint}, where we measure localization error as a function of angular separation between image pairs. All baseline methods experience steadily increasing error as the viewpoint gap grows, consistent with the observation from~\cite{banani2024probing} that single-view features struggle under large viewpoint changes. In contrast, our method exhibits a remarkably flat curve, maintaining low error even at extreme angles - a strong indicator of genuine 3D-awareness rather than superficial feature matching.

\begin{figure*}[t]
    \centering
    \includegraphics[width=0.95\linewidth]{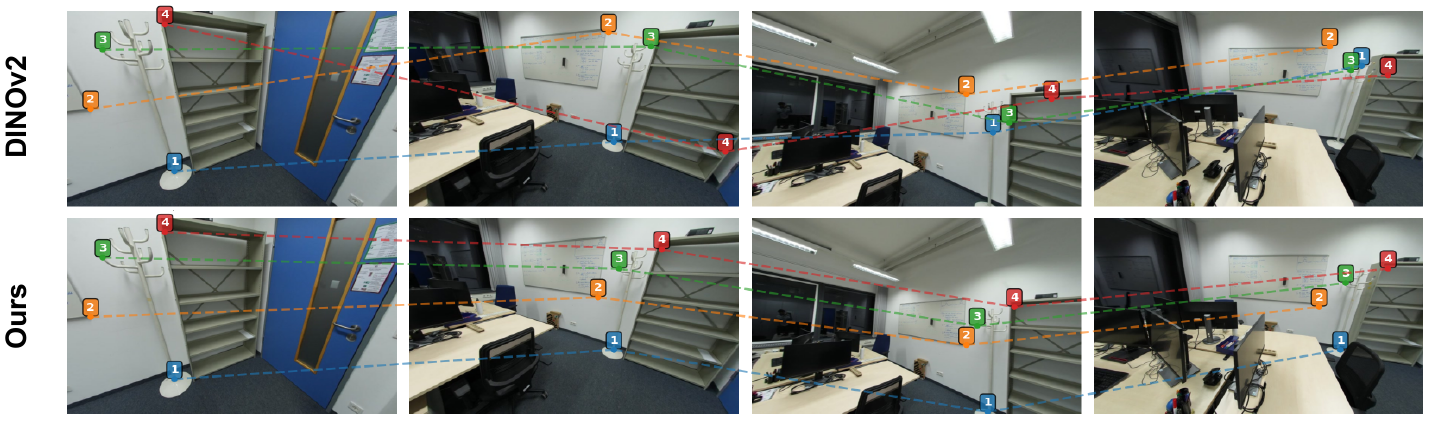}
    \caption{
    \textbf{Feature Consistency Across Views.} Numbered markers indicate query points in the first image, with dashed lines connecting to the most similar features in other views. Our method maintains geometric consistency with correspondences converging to the same 3D locations, while the original DINOv2 exhibits geometric drift across viewpoints.    
    }
    \label{fig:corr_example}
\end{figure*}
Figure~\ref{fig:corr_example} provides a visual illustration: given the same query points, DINOv2 produces correspondences that drift significantly across views, while our method maintains accurate alignment. Importantly, these correspondence metrics are not the end product - they serve as a rigorous proof of multi-view consistency in the learned feature space.

\begin{wraptable}{r}{0.4\linewidth}
\centering
\vspace{-12pt}
\caption{
\textbf{3D-Aligned Surface Normal Estimation}. Percentage recall at different thresholds and the root-mean square angular prediction error between predicted normals and the ground truth on the Navi dataset. 
}
\footnotesize 
\setlength{\tabcolsep}{3pt} 
\resizebox{\linewidth}{!}{%
\begin{tabular}{lcccc}
\toprule
\textbf{Model} & \textbf{$11.25^{\circ} \uparrow$} & \textbf{$22.5^{\circ} \uparrow$} & \textbf{$30^{\circ} \uparrow$} & \textbf{RMSE $\downarrow$}\\
\midrule
DINOv2 & 0.156 & 0.336 & 0.434  & 56.00 \\
MEF & 0.149 & 0.338 & 0.440 & 54.77 \\
FiT3D & 0.176 & 0.385 & 0.494  & 51.00 \\
SnD & 0.168 & 0.359 & 0.460 & 53.61 \\
\midrule
\textbf{Ours} & \textbf{0.251} & \textbf{0.553} & \textbf{0.693} & \textbf{32.17} \\
\bottomrule
\end{tabular}
}
\label{tab:snorm}
\end{wraptable}
\paragraph{\textbf{3D Aligned Normal Estimation.}} 
For surface normal estimation, following Banani~\etal\cite{banani2024probing}, we freeze all backbone weights and train a lightweight probe to predict normals. However, we extend their single-view setup in two ways: (1)~we use a multi-view probe that aggregates information across viewpoints via self-attention, and (2)~rather than predicting normals in each camera's local coordinate system, we require the probe to output normals aligned to a single shared 3D reference frame. This formulation directly tests whether the features encode a coherent, view-consistent 3D structure.

Table~\ref{tab:snorm} shows that our method outperforms both base DINOv2 and other lifting methods by a wide margin across all angular thresholds and RMSE. Critically, the backbone remains frozen during probing - the probe does not learn 3D reasoning, but merely reads out the geometric structure already present in the feature space. The substantial gap over other lifting methods, which also target geometric improvement, suggests that our multi-view processing encodes fundamentally richer 3D information than single-view lifting approaches.

Figure~\ref{fig:snorm} illustrates this qualitatively: our predicted normals closely match the ground truth, producing smooth and view-consistent surfaces, while DINOv2 and FiT3D exhibit noticeable artifacts and inconsistencies across viewpoints. In particular, surfaces facing the same world direction retain consistent normal orientation regardless of camera angle - a direct consequence of the 3D-aware structure encoded in our features.

\begin{figure*}[t]
    \centering 
    
    \setlength{\tabcolsep}{4pt} 
    
    \begin{tabular}{cccccc} 
        \textbf{Input Images} & \textbf{Ground Truth} & \textbf{Ours} & \textbf{DINOv2} & \textbf{FiT3D} \\

        \includegraphics[width=0.18\textwidth]{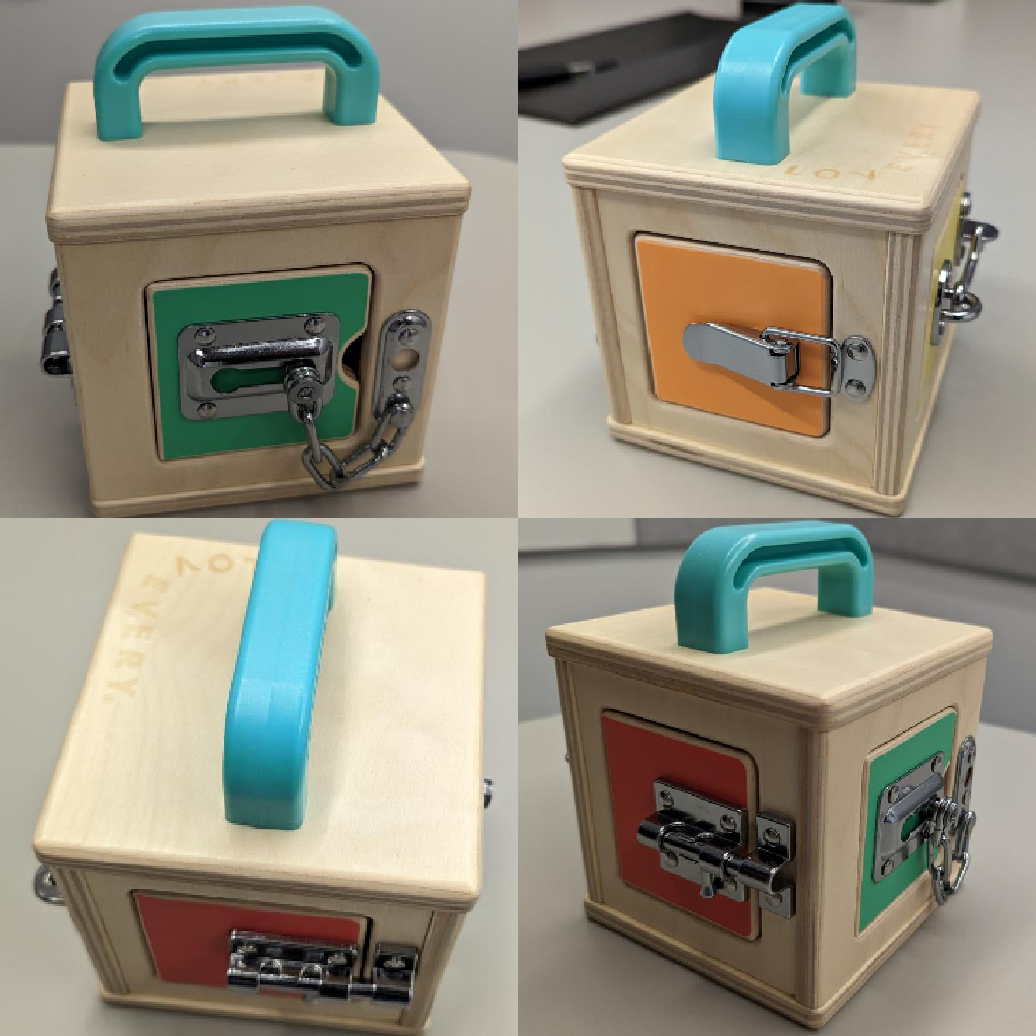} &
        \includegraphics[width=0.18\textwidth]{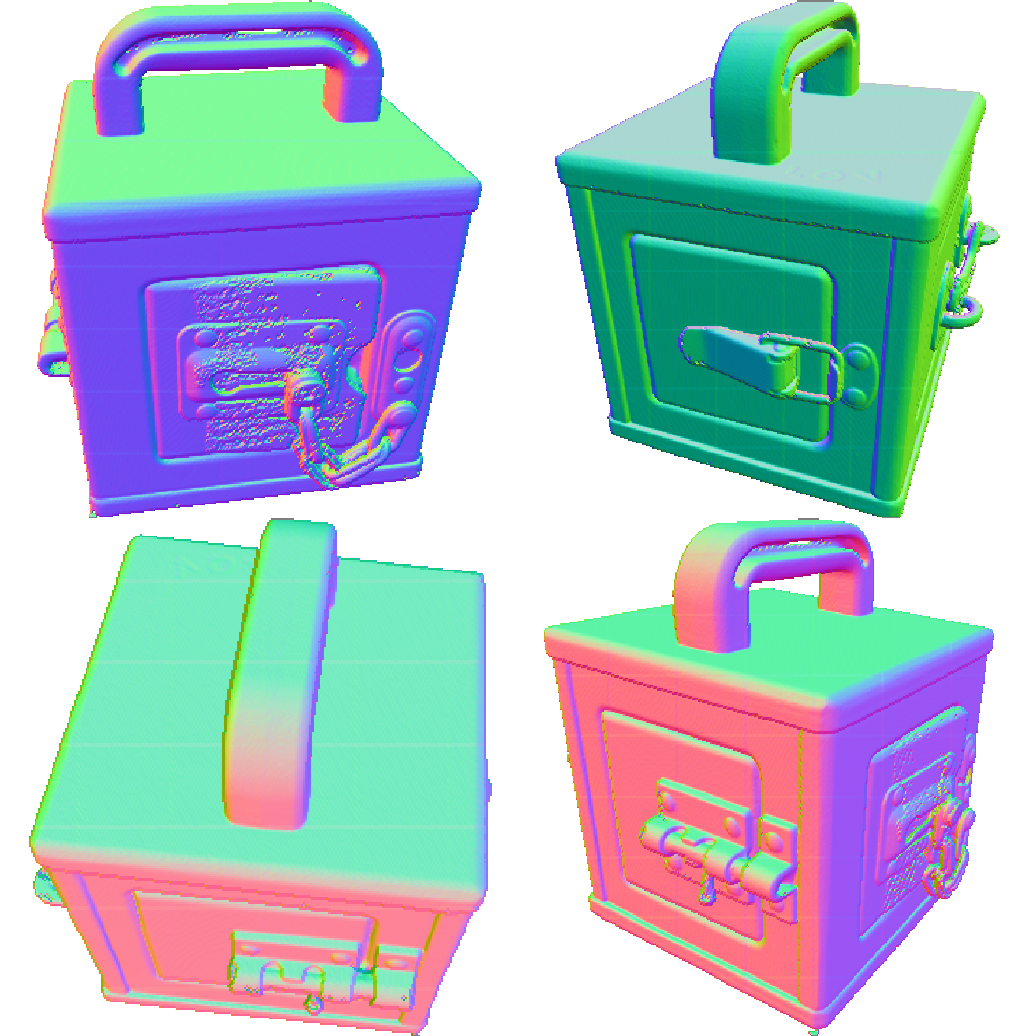} &
        \includegraphics[width=0.18\textwidth]{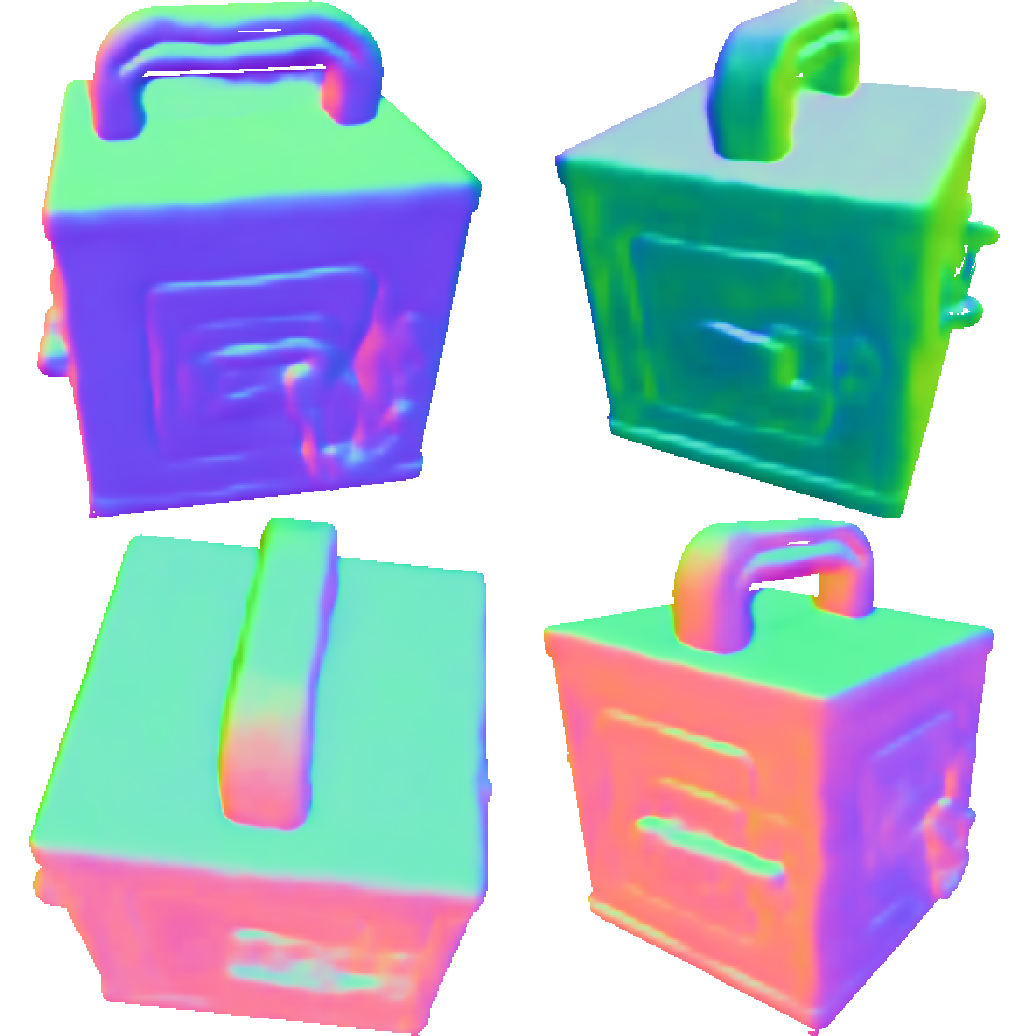} &
        \includegraphics[width=0.18\textwidth]{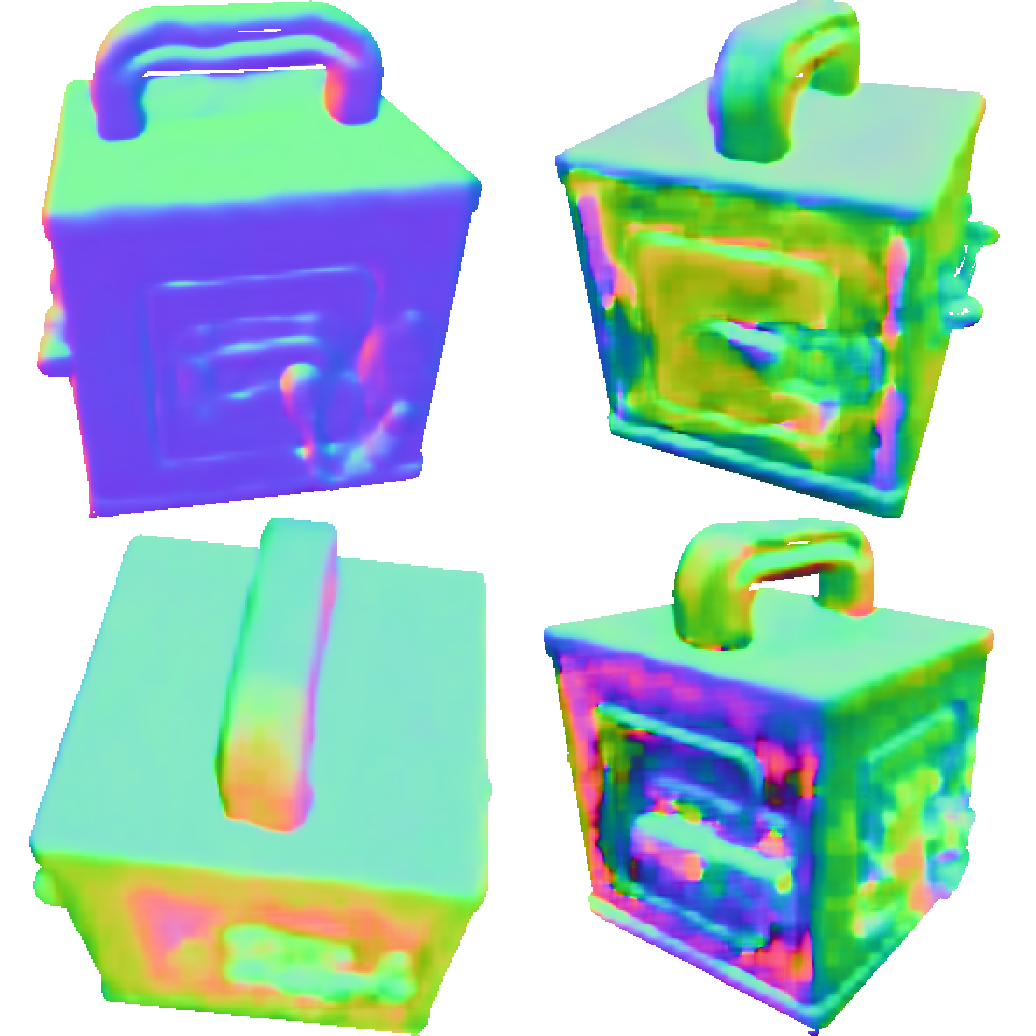} &
        \includegraphics[width=0.18\textwidth]{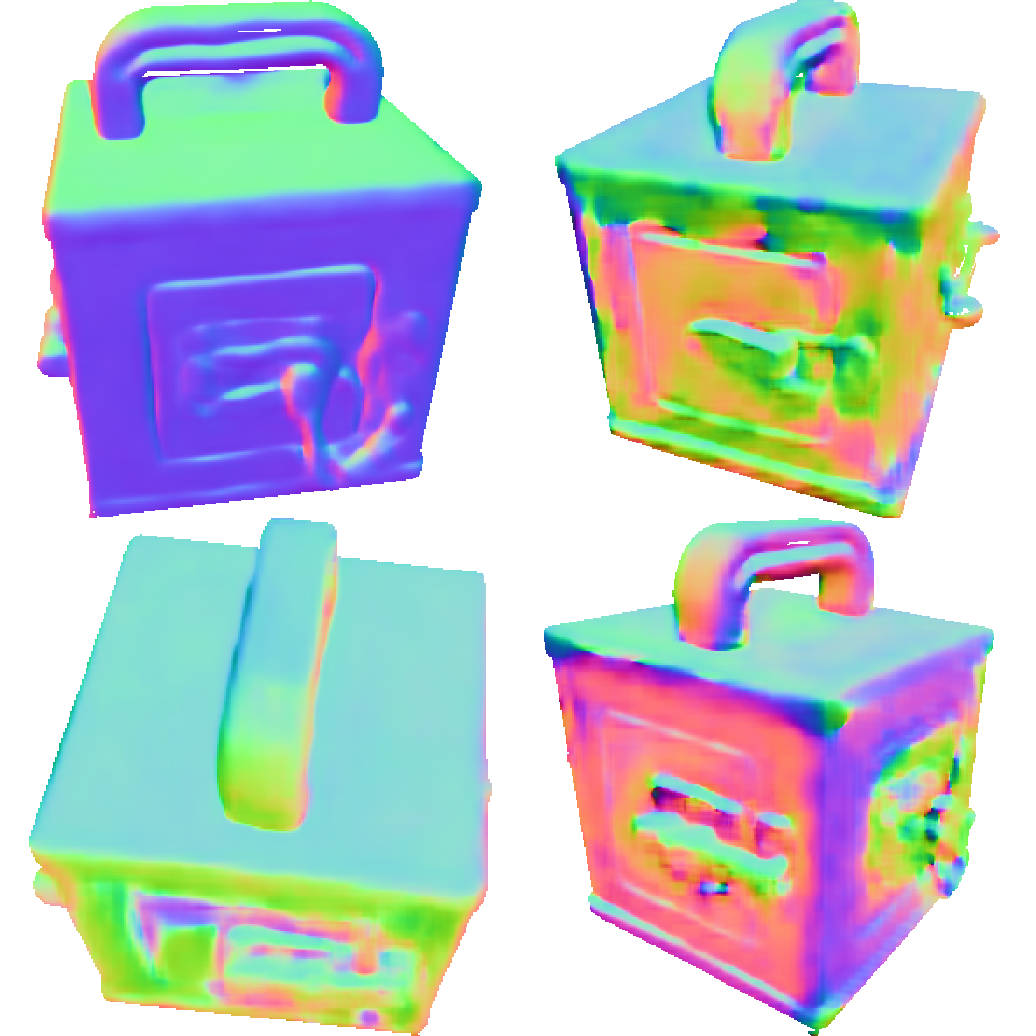}  \\

        
        
    \end{tabular}
    \caption{\textbf{Qualitative Results for 3D-Aligned Surface Normal Estimation.} We visualize the predicted surface normals from multiple viewpoints. The consistency in color and orientation across different views demonstrates that our multi-view features encode a coherent 3D representation. Note how surfaces facing the same world direction retain the same color regardless of the camera angle. 
    }
    \label{fig:snorm}

\end{figure*}

\subsection{Generalization Across Foundation Models}
A key design goal of our framework is generality - the ability to lift any transformer-based foundation model into 3D without architectural modifications or costly per-model data pipelines (e.g., reconstructing a Gaussian Splatting scene for each feature space). To demonstrate this, we apply our method to three additional backbones beyond DINOv2: CLIP~\cite{radford2021learning}, DINOv3, and SAM~\cite{kirillov2023segment}. As shown in Table~\ref{tab:generalization_models}, our method consistently reduces localization error across all models and benchmarks while maintaining high similarity to the respective original foundation model. Notably, even models that start with substantially higher baseline error, such as SAM and CLIP, benefit from the same training recipe and achieve comparable 3D-awareness to our DINOv2 results. This demonstrates that the geometric signal learned by our spatial adapters is not specific to any particular feature space, but rather captures a general multi-view structure that transfers across foundation models. We show additional metrics and comparisons in the supplementary.
\begin{table*}[t]
\centering
\caption{
\textbf{Generalization Across Foundation Models.}
Our framework generalizes to multiple foundation models without architectural changes. We report correspondence localization error on ScanNet++ and generalization benchmarks, and cosine similarity to the respective base model.
}
\resizebox{\textwidth}{!}{%
\begin{tabular}{l|cc|cc|cc|cc|cc|cc}
\toprule
& \multicolumn{4}{c|}{\textbf{CLIP}} & \multicolumn{4}{c|}{\textbf{DINOv3}} & \multicolumn{4}{c}{\textbf{SAM}} \\
\cmidrule(lr){2-5} \cmidrule(lr){6-9} \cmidrule(lr){10-13}
& \multicolumn{2}{c}{ScanNet++} & \multicolumn{2}{c|}{General.}
& \multicolumn{2}{c}{ScanNet++} & \multicolumn{2}{c|}{General.}
& \multicolumn{2}{c}{ScanNet++} & \multicolumn{2}{c}{General.} \\
\cmidrule(lr){2-3} \cmidrule(lr){4-5} \cmidrule(lr){6-7} \cmidrule(lr){8-9} \cmidrule(lr){10-11} \cmidrule(lr){12-13}
& Loc.\ Err.~$\downarrow$ & Sim.~$\uparrow$
& Loc.\ Err.~$\downarrow$ & Sim.~$\uparrow$
& Loc.\ Err.~$\downarrow$ & Sim.~$\uparrow$
& Loc.\ Err.~$\downarrow$ & Sim.~$\uparrow$
& Loc.\ Err.~$\downarrow$ & Sim.~$\uparrow$
& Loc.\ Err.~$\downarrow$ & Sim.~$\uparrow$ \\
\midrule
Base
& 0.1505 & --
& 0.1937 & --
& 0.1025 & --
& 0.1302 & --
& 0.1846 & --
& 0.1833 & -- \\
\textbf{Ours}
& \textbf{0.0301} & \textbf{0.8824}
& \textbf{0.0881} & \textbf{0.8389}
& \textbf{0.0314} & \textbf{0.9411}
& \textbf{0.1027} & \textbf{0.9179}
& \textbf{0.0244} & \textbf{0.9547}
& \textbf{0.0752} & \textbf{0.9488} \\
\bottomrule
\end{tabular}%
}
\label{tab:generalization_models}
\end{table*}
\begin{figure}
    \centering
    \includegraphics[width=0.8\linewidth]{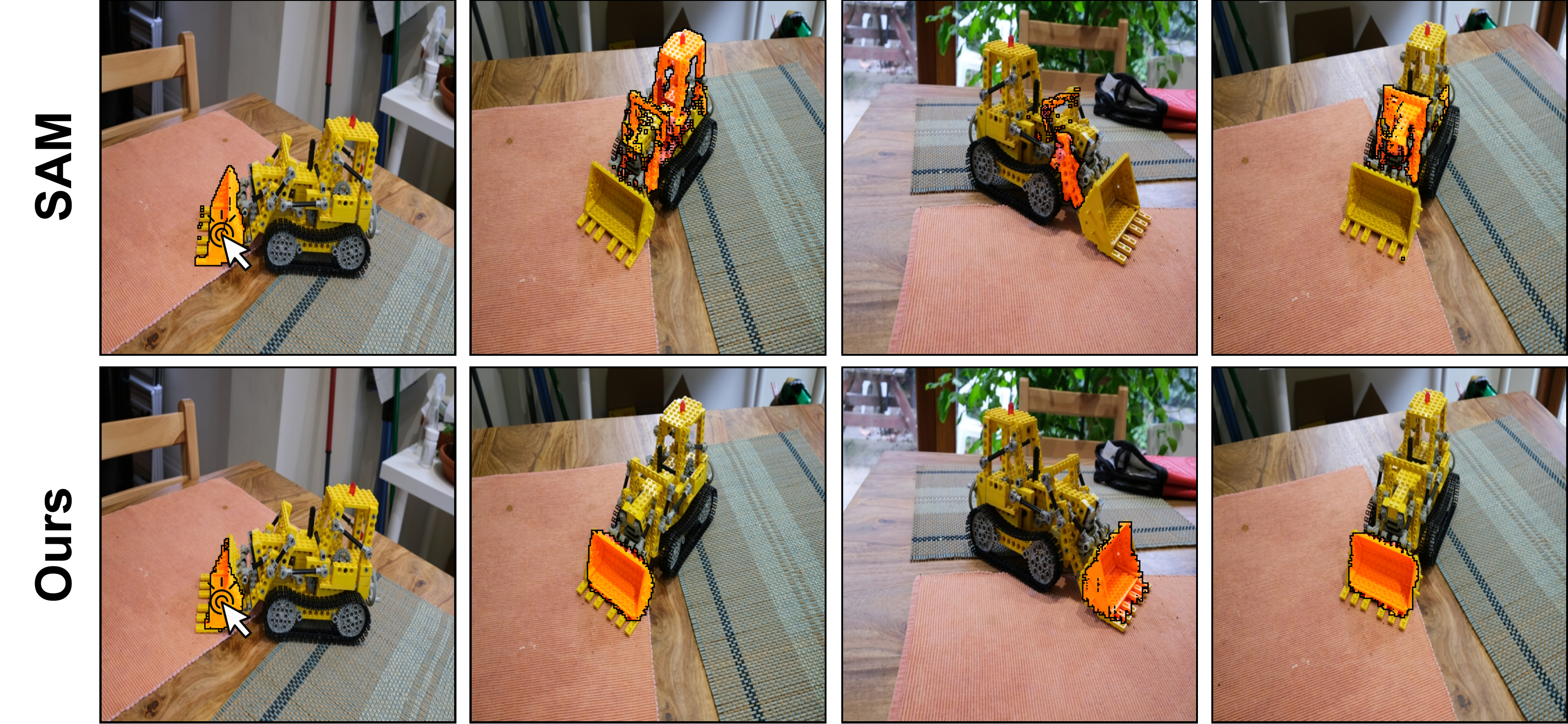}
    \caption{
    \textbf{Consistent Multi-View Segmentation.} 
    A single click in the first view instantly yields consistent semantic segmentation masks across all views using our method, whereas standard SAM fails to maintain such consistency. 
    }
    \label{fig:sam}
\end{figure}

\paragraph{\textbf{Cross-View Semantic Segmentation.}}
We further illustrate this generality with a practical example using SAM. Figure~\ref{fig:sam} demonstrates that our method enables consistent segmentation across multiple views from a single user click, while the original SAM fails to maintain correspondence. Given the feature at the user's click, we retrieve its nearest match in other views and use the resulting locations as SAM prompts - the same correspondence mechanism used in our geometric evaluation, now directly enabling a multi-view application. Our model already achieves consistent cross-view segmentation without any additional training. Results are further improved by briefly fine-tuning the SAM decoder and prompt encoder (with the backbone frozen) using self-supervision from the original model's outputs - a lightweight adaptation that requires no ground-truth labels. The resulting consistency across wide viewpoint changes confirms that our framework preserves both geometric alignment and semantic coherence.

\subsection{Ablation Study}
We analyze the contribution of individual design choices in Table~\ref{ablation}. Our correspondence training objective alone improves multi-view consistency but suffers from high location error and low semantic similarity when limited to single-view processing. Adding the multi-view adapter substantially reduces location error and improves semantic alignment by enabling cross-view reasoning. The Plücker embedding provides further gains by explicitly encoding camera geometry. Notably, our method achieves impressive multi-view consistency even without camera information. This offers users a valuable trade-off, enabling the framework's use when camera data is unavailable, or allowing for further performance gains when it is. Finally, removing regularization yields the best geometric consistency but at the cost of significant semantic drift from the base model - our final design balances both objectives.

In the supplementary, we include extensive evaluations and additional experiments, such as analyses of correspondence accuracy as the number of input views increases, measurements of inference time as a function of the number of input views, and experiments isolating the effects of regularization strength, context stability, and correspondence density on the learned multi-view representations.


\begin{table}[t]
\centering
\small
\caption{
\textbf{Design Ablation}. We showcase the contribution of each of our suggested modifications and losses.}
\begin{tabular}{lcc}
\toprule
\textbf{Model} & \textbf{Loc.~Err.~(↓)} & \textbf{Base~Sim.~(↑)} \\
\midrule
Base DINOv2 & 0.1029 & - \\
(+) Lora Training & 0.0347 & 0.9117 \\
(+) Multi-View Adapter  & 0.0262 & 0.9360 \\
(+) \textbf{Plücker (Full method)} & 0.0247 & 0.9376 \\
\midrule
(-) Regularization & 0.0204 & 0.1054 \\
\bottomrule
\end{tabular}

\label{ablation}
\end{table}

\section{Conclusions}


In this work, we address the inherent "3D blindness" of 2D foundation models by introducing a framework for joint multi-view processing. Unlike traditional "lifting" schemes that process images in isolation, our approach "inflates" existing backbones to natively handle multi-view inputs. By interleaving multi-view adapters directly within the pre-trained layers, we facilitate seamless cross-view communication while strictly preserving the original semantic manifold.

By bypassing computationally expensive per-scene optimizations (such as 3DGS) in favor of a training pipeline based on geometric correspondences, we provide an efficient path for adapting any new foundation model. 
Our evaluations across DINOv2, DINOv3, CLIP, and SAM demonstrate that our method significantly enhances 3D awareness without sacrificing the zero-shot capabilities that make these models so valuable.
Our approach offers a scalable and effective blueprint for bridging the gap between 2D semantic intelligence and 3D spatial coherence. We believe this shift toward joint, multi-view aware architectures is a crucial step toward creating foundation models that truly understand the three-dimensional nature of our world.

\paragraph{Acknowledgment:} Part of this research was supported by ISF grant 2132/23.



%
%
\bibliographystyle{splncs04}
\bibliography{main}

\section{Appendix Overview}
This supplementary provides additional qualitative and quantitative results supporting the main paper. Specifically, we include:
\begin{enumerate}
    \item \textbf{Additional Ablation Studies} - We include an extensive ablation, including analyses of correspondence accuracy as the number of input views increases, measurements of inference time as a function of the number of input views, experiments isolating the effects of regularization strength, context stability, and correspondence density on the learned multi-view representations.
    \item \textbf{Additional Results} - extended visualizations of feature-space comparisons and 3D-aligned surface-normal predictions.
    \item \textbf{Implementation Details} - full technical specifications of training settings, architectural choices, and the multi-view probing setup.
\end{enumerate}

\section{Additional Ablation Study}

\paragraph{Scaling with Number of Views}
To understand how our method's performance scales with the number of views, we conduct an ablation study on correspondence prediction between two query views. Starting with just the query pair (2 views), we progressively add context views using two strategies: (1) \textbf{Random}: uniformly sampling views from the scene, and (2) \textbf{Meaningful}: selecting views spatially close to the query pair based on camera positions.

\begin{figure}
    \centering
    \includegraphics[width=0.8\linewidth]{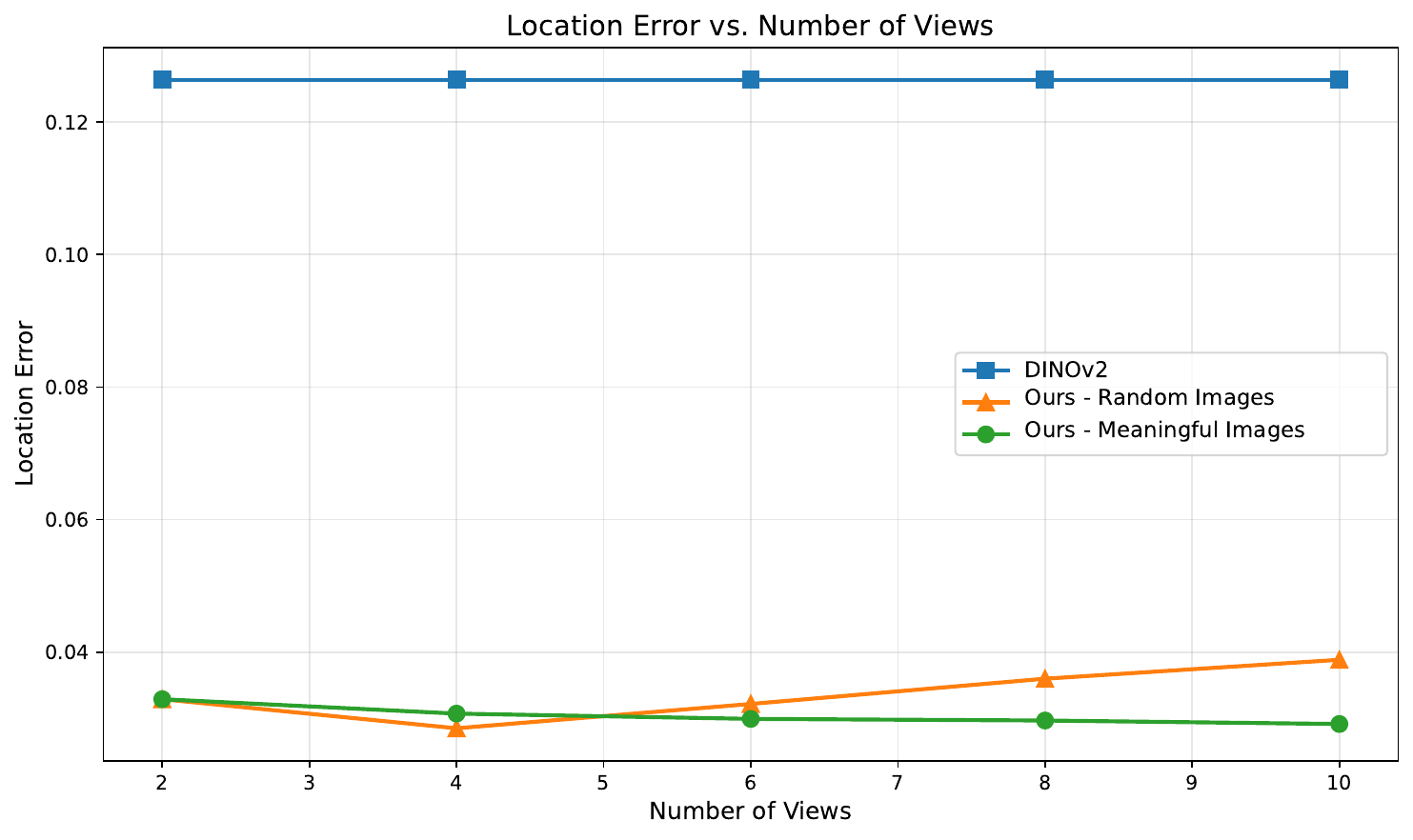}
    \caption{\textbf{Multi-view scalability analysis.} Our method achieves~4× lower location error than the original DINOv2 for correspondence prediction. At 2 views (the query pair only), our method demonstrates the value of multi-view reasoning. Adding meaningful context views (nearby images) provides further improvement, while random context views degrade performance at higher view counts. Evaluated on the ScanNet++ test split.}
    \label{fig:num_views_error}
\end{figure}

We measure location error for correspondences between the fixed query pair while varying the total number of views from 2 to 10. Experiments are performed on the ScanNet++ test split. As shown in Figure \ref{fig:num_views_error}, our method achieves approximately 4× lower location error than baseline DINOv2 (~0.033 vs. ~0.13) already at 2 views, demonstrating the substantial benefit of multi-view reasoning. Importantly, this improvement comes primarily from our training strategy and our multi-view attention between the two query images, enabling these images to attend to each other rather than being processed independently.
Additional context views provide incremental benefits when selected appropriately. our method with meaningful context views shows continued improvement as more views are added, declining from 0.033 at 2 views to 0.029 at 10 views. In contrast, random context views exhibit a U-shaped pattern: performance initially improves to 0.028 at 4 views but gradually degrades beyond 6 views, reaching 0.039 at 10 views. This degradation suggests that spatially distant or irrelevant views introduce noise into the multi-view attention mechanism.
These results demonstrate that while the core multi-view reasoning between the query pair provides the primary performance gain, carefully selected context views can further enhance correspondence quality.

\begin{figure}
    \centering
    \includegraphics[width=0.8\linewidth]{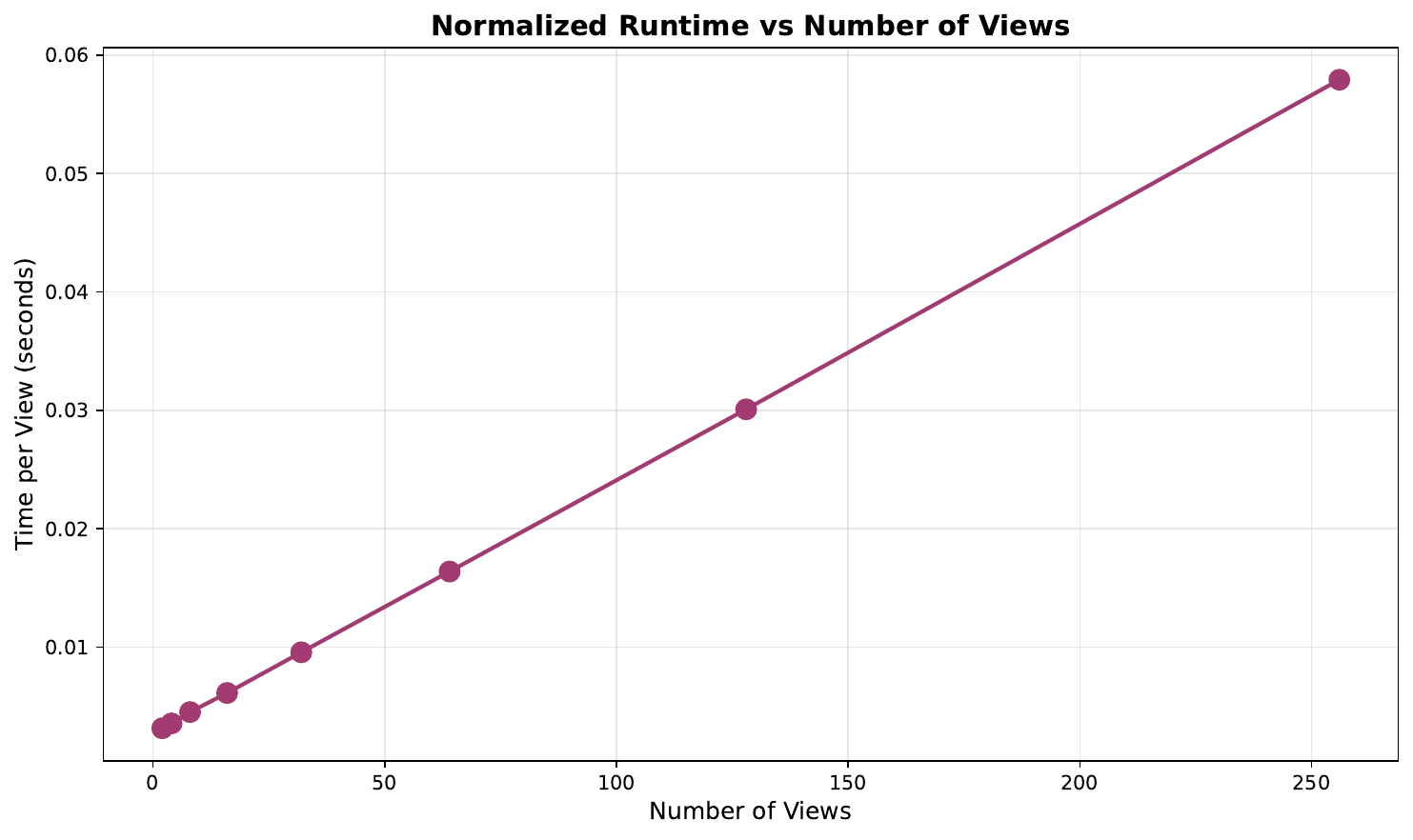}
    \caption{\textbf{Runtime Analysis.} Normalized runtime per view as a function of the number of views. The processing time per view increases linearly with the total number of views.}
    \label{fig:runtime}
\end{figure}

\paragraph{Feed-Forward Speed Analysis}
The results demonstrate a linear increase in processing time per view as the number of views grows, which is directly attributable to the multi-view attention mechanism. As shown in Table~\ref{tab:time_per_view} and Figure~\ref{fig:runtime}, the time per view increases from 0.00315 seconds for 2 views to 0.05793 seconds for 256 views, with a slope of approximately $2.16 \times 10^{-4}$ seconds per additional view. This linear scaling is expected behavior for optimized attention, where each view must attend to all other views in the set. The computational complexity increases with the number of pairwise interactions, resulting in $O(n)$ growth in per-view processing time as the total number of views $n$ increases. Despite this linear growth, the method remains efficient even at 256 views, with per-view processing time under 60 milliseconds.

\begin{table}[h]
\centering
\tiny
\caption{\textbf{Runtime Analysis.} Processing time per view as a function of the number of views.}
\begin{tabular}{cc}
\toprule
Num Views & Time per View (sec) \\
\midrule
2 & 0.00315 \\
4 & 0.00356 \\
8 & 0.00452 \\
16 & 0.00613 \\
32 & 0.00957 \\
64 & 0.01639 \\
128 & 0.03009 \\
256 & 0.05793 \\
\bottomrule
\end{tabular}
\label{tab:time_per_view}
\end{table}

\paragraph{Error Consistency.}
Our method builds upon a 3D attention mechanism, where each image feature attends not only to its own spatial tokens but also to features from other views in the set, enriched with their corresponding Plücker embeddings.
As a result, the representation of a single image can depend on the other views provided as context.
To assess the stability of this dependency, we perform an error consistency ablation that measures how changing the contextual images affects the predicted correspondences.

In this experiment, we fix a pair of reference images and repeatedly replace the remaining two images in the multi-view set with random ones.
We then compute the location error between the fixed image pair for each configuration, evaluating how much the correspondence accuracy fluctuates with different contextual inputs.
We perform this experiment on 50 fixed image pairs, each tested with 40 random context variations.
The model used in this experiment is based on the DINOv2 backbone.

As shown in \Cref{error_consistency_ablation}, the mean location error for our model remains significantly lower than that of the base model (0.0506 vs.\ 0.1017), and the variance across different context configurations is minimal (std.\ 0.0020).
This demonstrates that although our features adapt to the contextual views, the resulting correspondences remain highly stable, confirming that the 3D attention mechanism produces consistent and robust 3D-aware features.

\begin{table}[t]
\centering
\caption{\textbf{Error consistency ablation}. For each of 50 image pairs, we randomly replace the other two images with 40 alternatives and measure the location error (2{,}000 evaluations in total).}
\begin{tabular}{lcc}
\toprule
\textbf{Model} & \textbf{Mean Loc. Error~(↓)} & \textbf{STD~(↓)} \\
\midrule
DINOv2 & 0.1017 & -- \\
Ours   & 0.0506 & 0.0020 \\
\bottomrule
\end{tabular}
\label{error_consistency_ablation}
\end{table}

\begin{figure}
    \centering
    \includegraphics[width=0.7\linewidth]{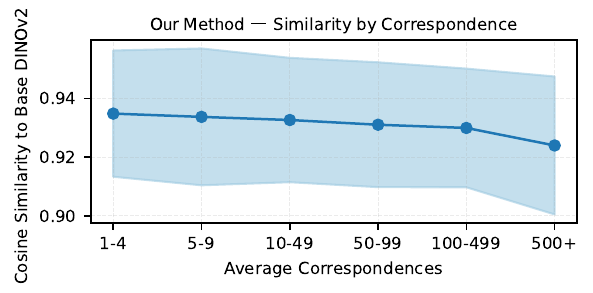}
    \caption{\textbf{Features Consistency.} 
     We visualize how the features change (compared to the base DINOv2 model) when the number of corresponding points between input views varies.
    }
    \label{fig:consistency}
\end{figure}

\paragraph{Feature Consistency.} 
We evaluate how the number of point correspondences between input images affects the resulting 3D-consistent features. When there is little overlap between input images, we expect our model’s output to remain close to that of the base foundation model. Conversely, when there is significant overlap, the output features should deviate from the base features to enforce consistency across corresponding regions. To analyze this behavior, we visualize a histogram showing how the number of correspondences between input images influences the similarity to the base foundation model’s features. Specifically, we use DINOv2 and measure feature similarity using cosine similarity. As shown in Figure~\ref{fig:consistency}, the output features are most similar to the base model when there is minimal overlap, while increasing the number of correspondences leads to greater deviation from the base model.

\begin{figure}
    \centering
    \includegraphics[width=0.5\linewidth]{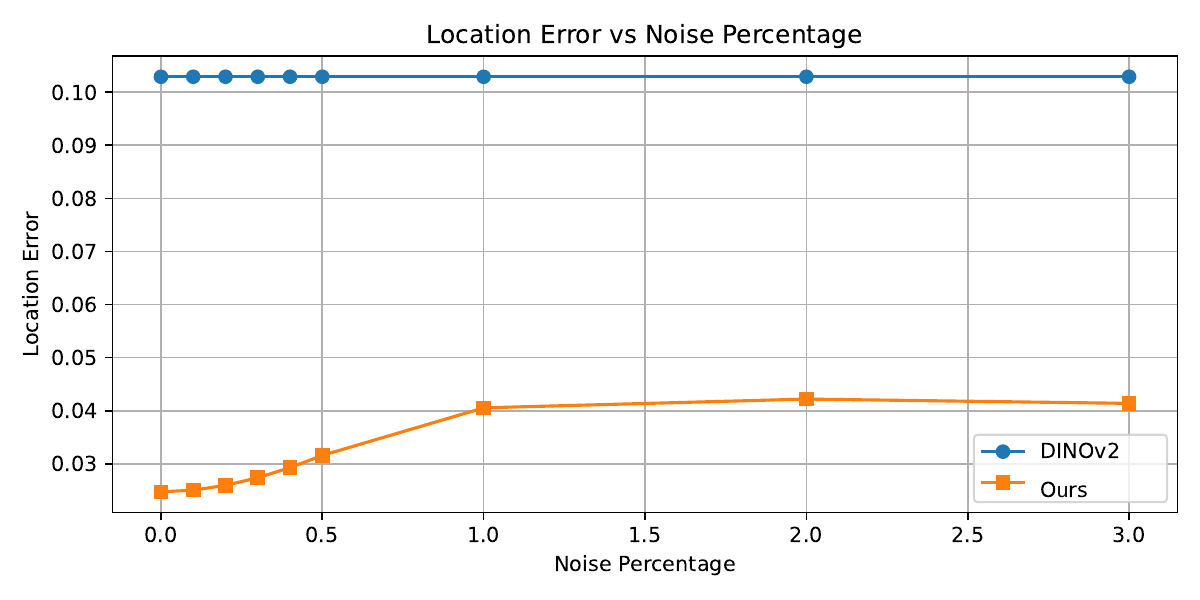}
    \caption{\textbf{Noisy Camera Parameters.} We evaluate the robustness of our method by perturbing camera extrinsics with increasing levels of Gaussian noise in both rotation and translation. The method remains stable under moderate amounts of noise, demonstrating tolerance to realistic calibration errors, while extreme perturbations lead to the expected drop in accuracy.
    }
    \label{fig:noise}
\end{figure}
\paragraph{Noisy Camera Parameters.}
To evaluate the robustness of our method to imperfect camera calibration, we conduct an ablation study by perturbing the extrinsic parameters with varying levels of synthetic noise. We inject zero-mean Gaussian noise into the camera poses, where the standard deviation is determined by our noise level parameter. For rotations, this standard deviation scales random perturbations in the axis-angle representation. For translations, the standard deviation is further scaled by the magnitude of the translation vector to ensure the noise is relative to the camera's distance. 
As shown in Figure~\ref{fig:noise}, despite the degradation in pose accuracy, our method remains stable under moderate noise, demonstrating that our approach tolerates realistic calibration errors. As expected, we observe a decline in accuracy when the injected noise becomes severe.

\begin{figure}[t]
    \centering
    \begin{tabular}{c}
        \includegraphics[width=0.65\linewidth]{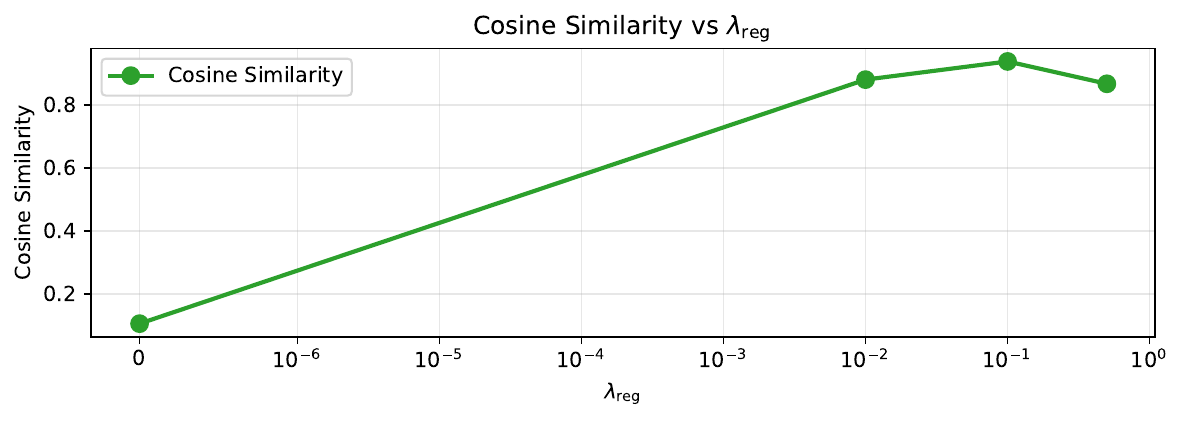} \\
        \includegraphics[width=0.65\linewidth]{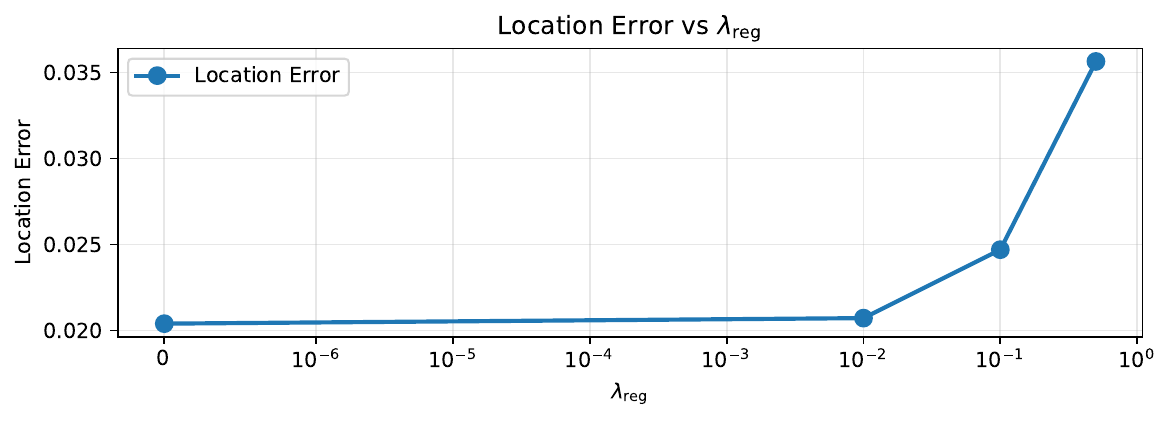} \\
    \end{tabular}
    \caption{\textbf{Ablation on $\lambda_\text{reg}$.} We show the trade-off between preserving the base-model feature space and minimizing localization error. Too little regularization distorts semantics, while too much limits learning of multi-view features.}
    \label{fig:ablation_lambda_reg}
\end{figure}
\paragraph{$\lambda_\text{reg}$ Ablation.}
Figure~\ref{fig:ablation_lambda_reg} presents the ablation results for the regularization weight $\lambda_\text{reg}$. The experiment highlights a clear trade-off between maintaining semantic consistency with the base model and achieving low localization error. When no regularization is applied, the model attains the lowest localization error but its feature space diverges noticeably from the original semantics. Conversely, applying strong regularization preserves the base-model structure but restricts the model’s ability to learn multi-view representations. These results indicate that a moderate level of regularization provides the best balance between accuracy and semantic stability.

\section{Additional Results}
\begin{figure}
    \centering
    \includegraphics[width=0.75\linewidth]{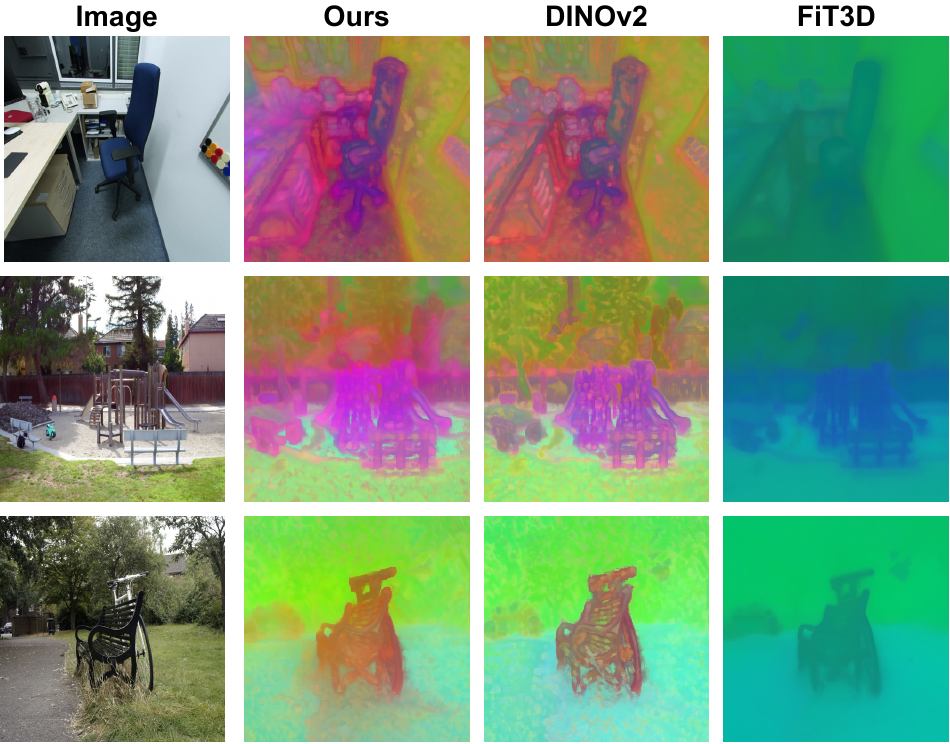}
\caption{
\textbf{Additional Feature Similarity Comparisons.} 
Shared PCA embeddings of our method, DINOv2, and FiT3D.
}    \label{fig:similarity_to_base_supp}
\end{figure}

Figure~\ref{fig:similarity_to_base_supp} presents additional examples comparing feature embeddings of our method, DINOv2, and FiT3D projected into a shared PCA space. Across diverse scenes, our method remains closely aligned with the base DINOv2 model, whereas FiT3D exhibits noticeably larger deviations, indicating reduced semantic fidelity. Complementary qualitative results in Figure~\ref{fig:snorm} visualize 3D-aligned surface-normal predictions from multiple viewpoints. The consistent colors and orientations across views - reflecting identical world-frame directions - demonstrate that our multi-view features encode a coherent and geometrically consistent 3D representation.

To further demonstrate practicality, we provide 3D point cloud reconstruction results (\Cref{tab:results_nocrop}). Coupled with the high base-model cosine similarity shown in the paper, we provide a robust upgrade for FM pipelines, enhancing multi-view consistency while preserving original semantics.
\begin{figure*}
    \centering 
    
    \setlength{\tabcolsep}{4pt} 
    
    \begin{tabular}{cccccc} 
        \textbf{Input Images} & \textbf{Ground Truth} & \textbf{Ours} & \textbf{DINOv2} & \textbf{FiT3D} \\

        \includegraphics[width=0.19\textwidth]{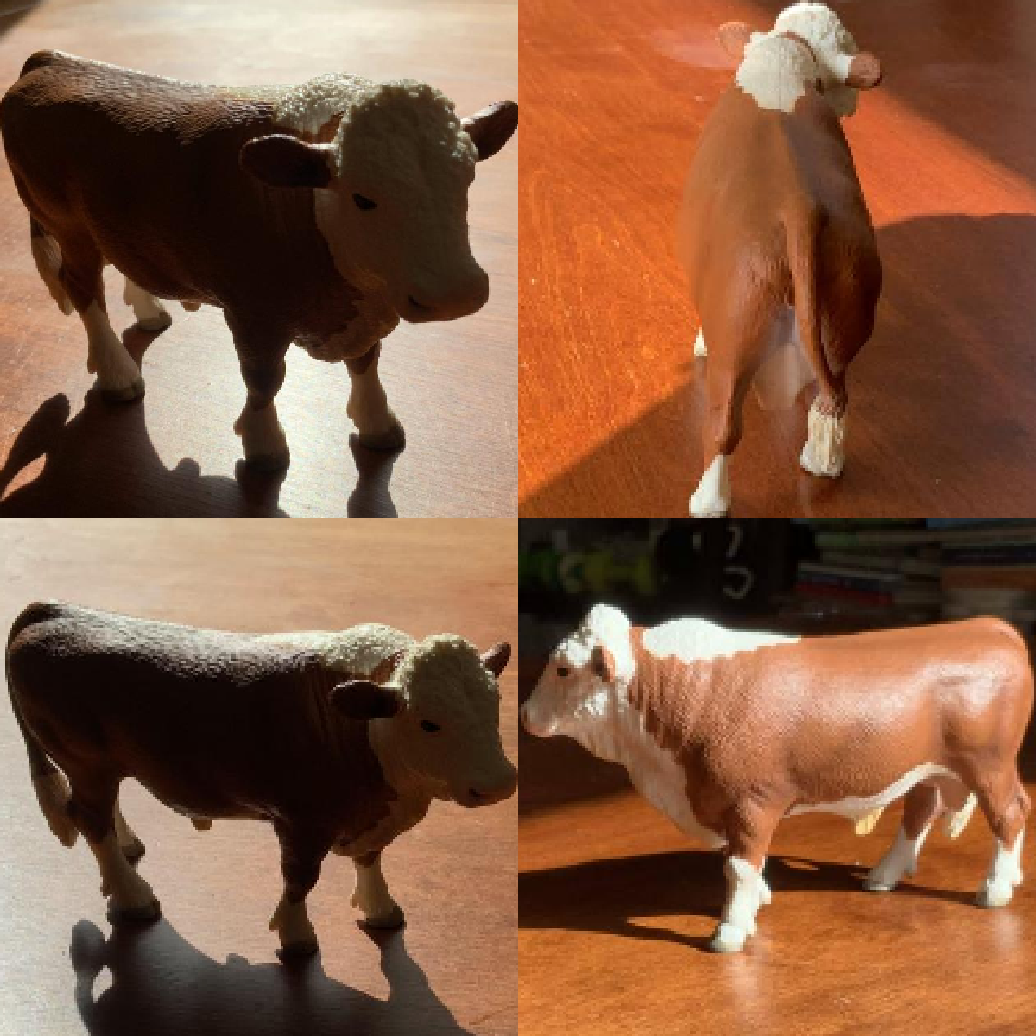} &
        \includegraphics[width=0.19\textwidth]{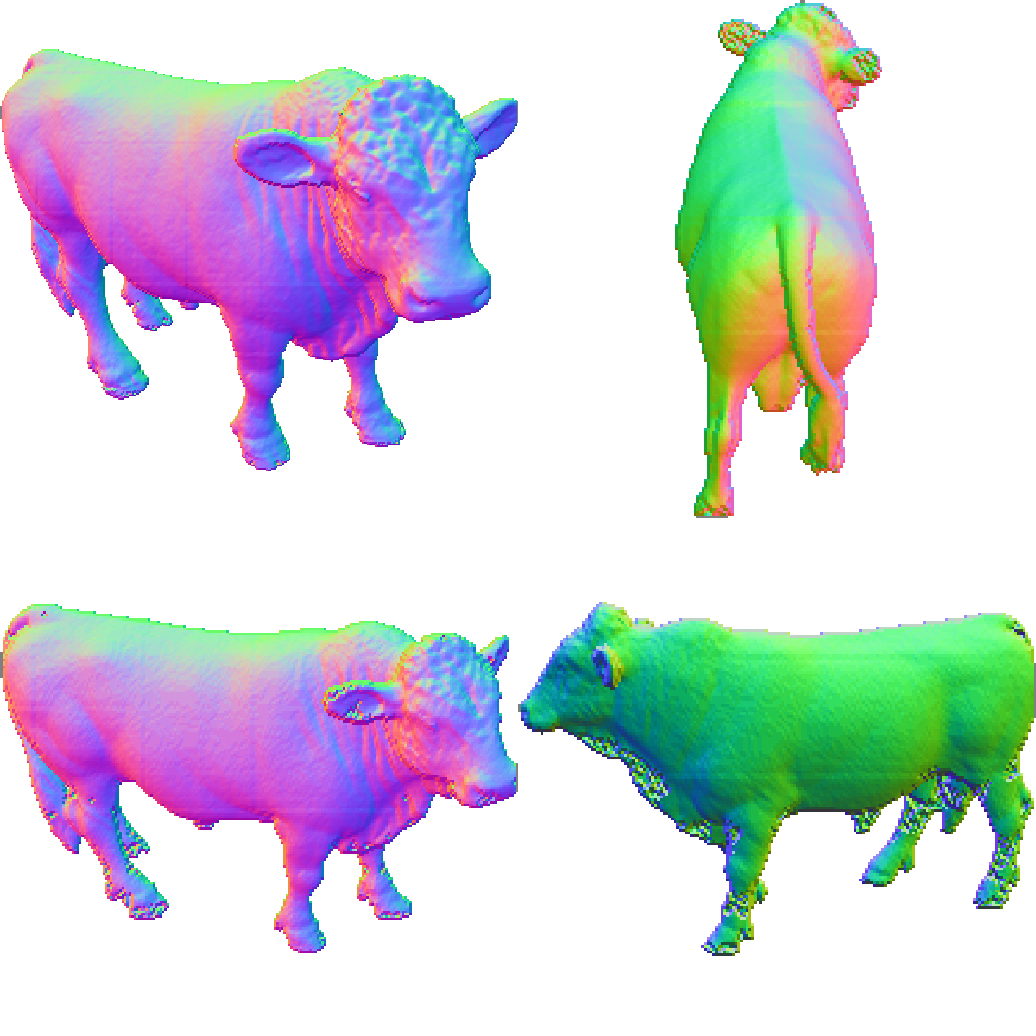} &
        \includegraphics[width=0.19\textwidth]{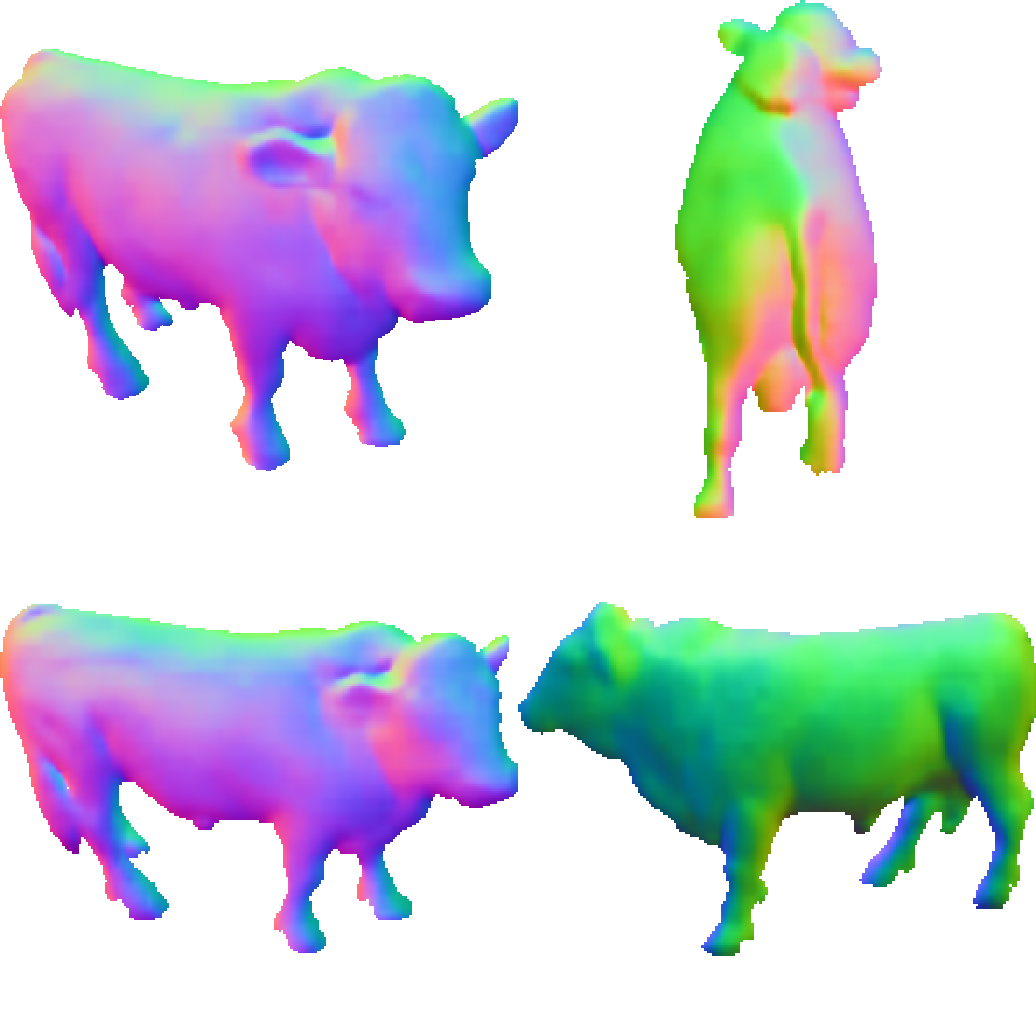} &
        \includegraphics[width=0.19\textwidth]{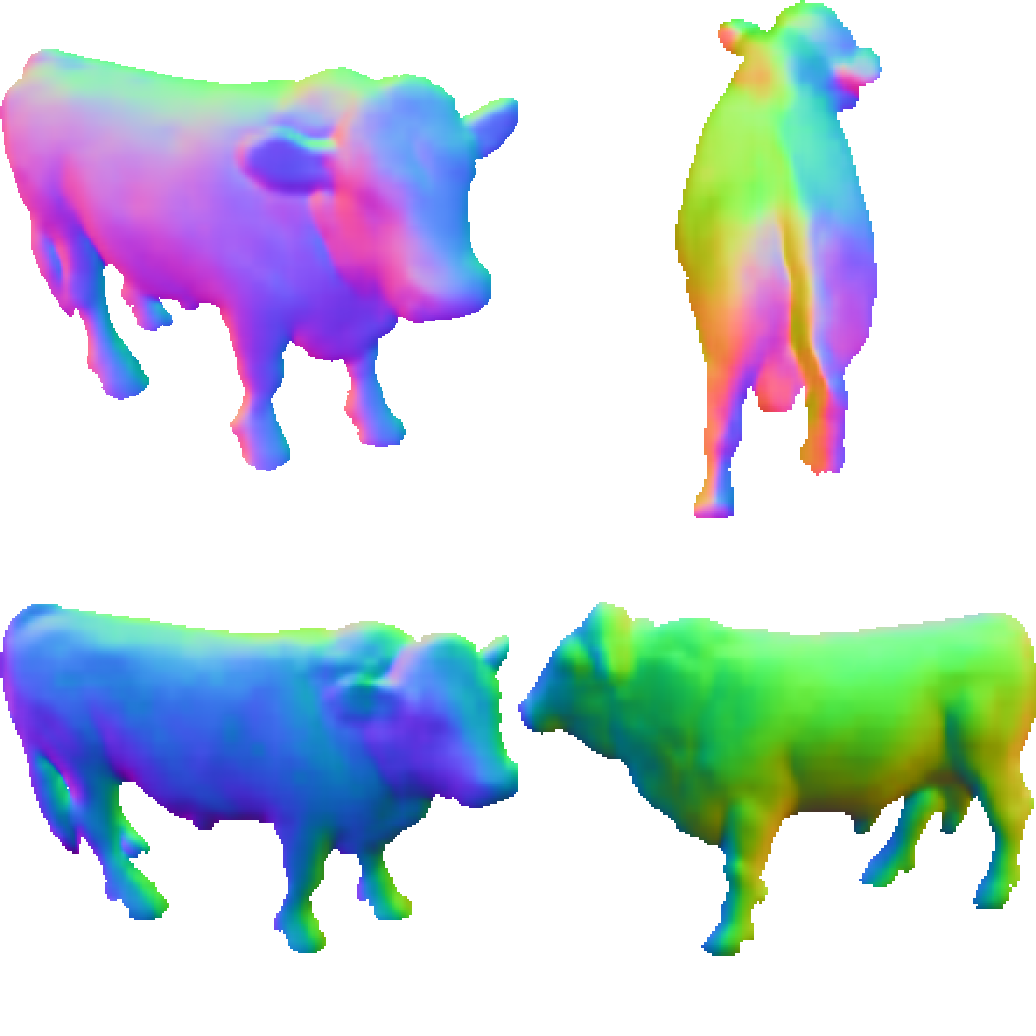} &
        \includegraphics[width=0.19\textwidth]{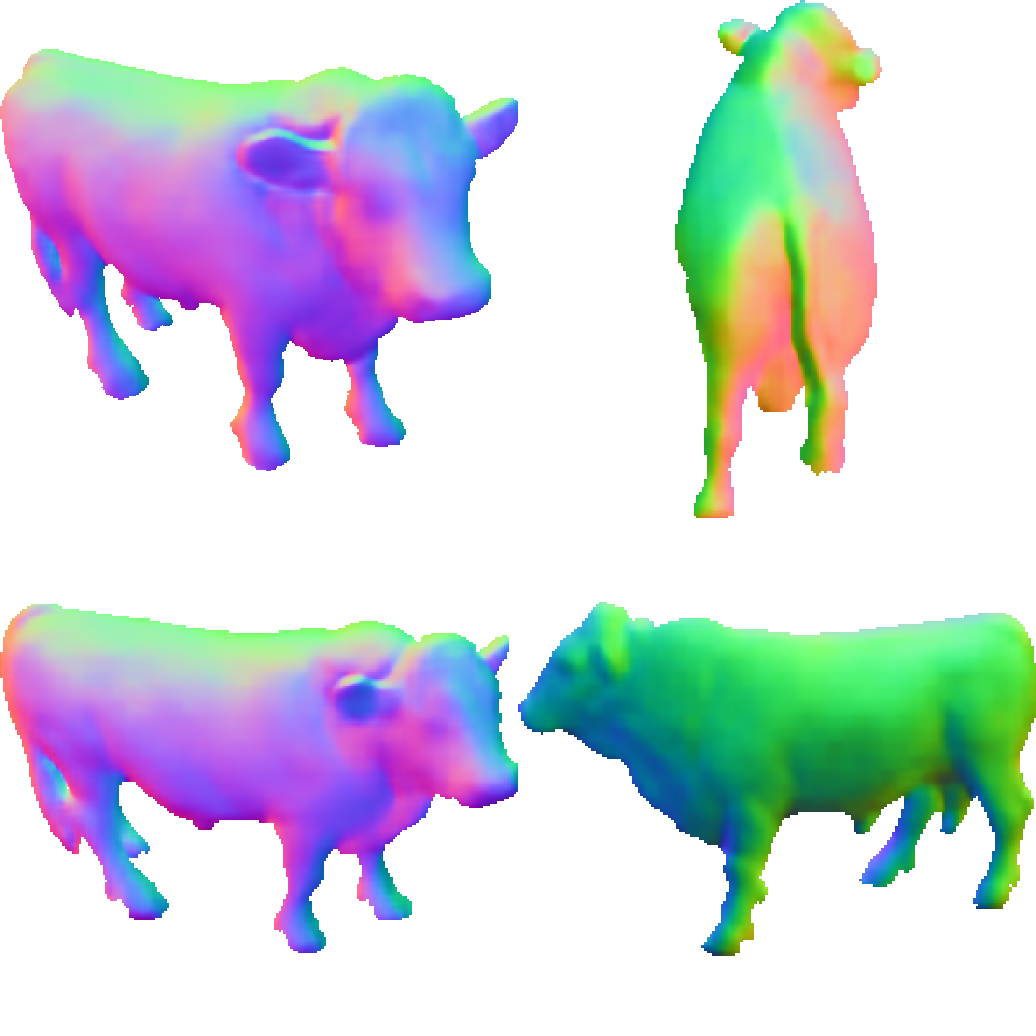} \\
        \includegraphics[width=0.19\textwidth]{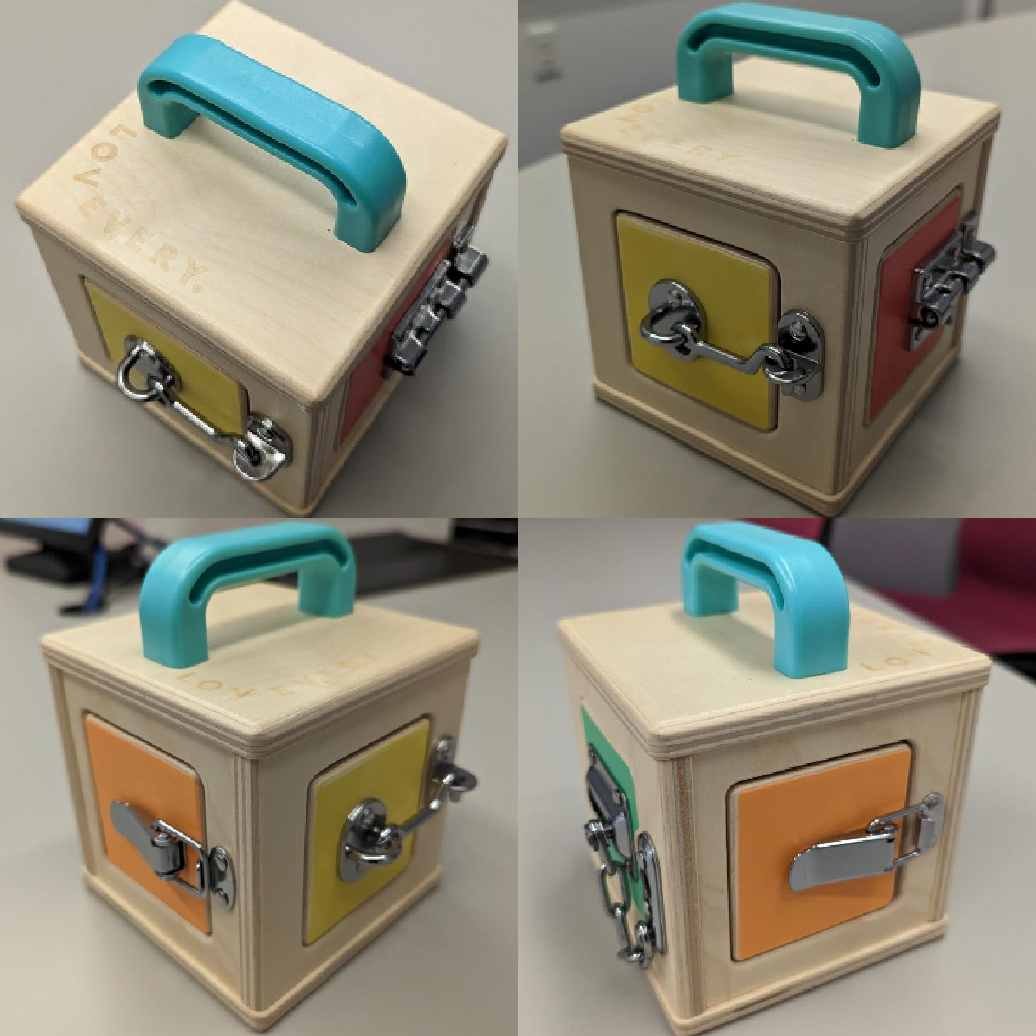} &
        \includegraphics[width=0.19\textwidth]{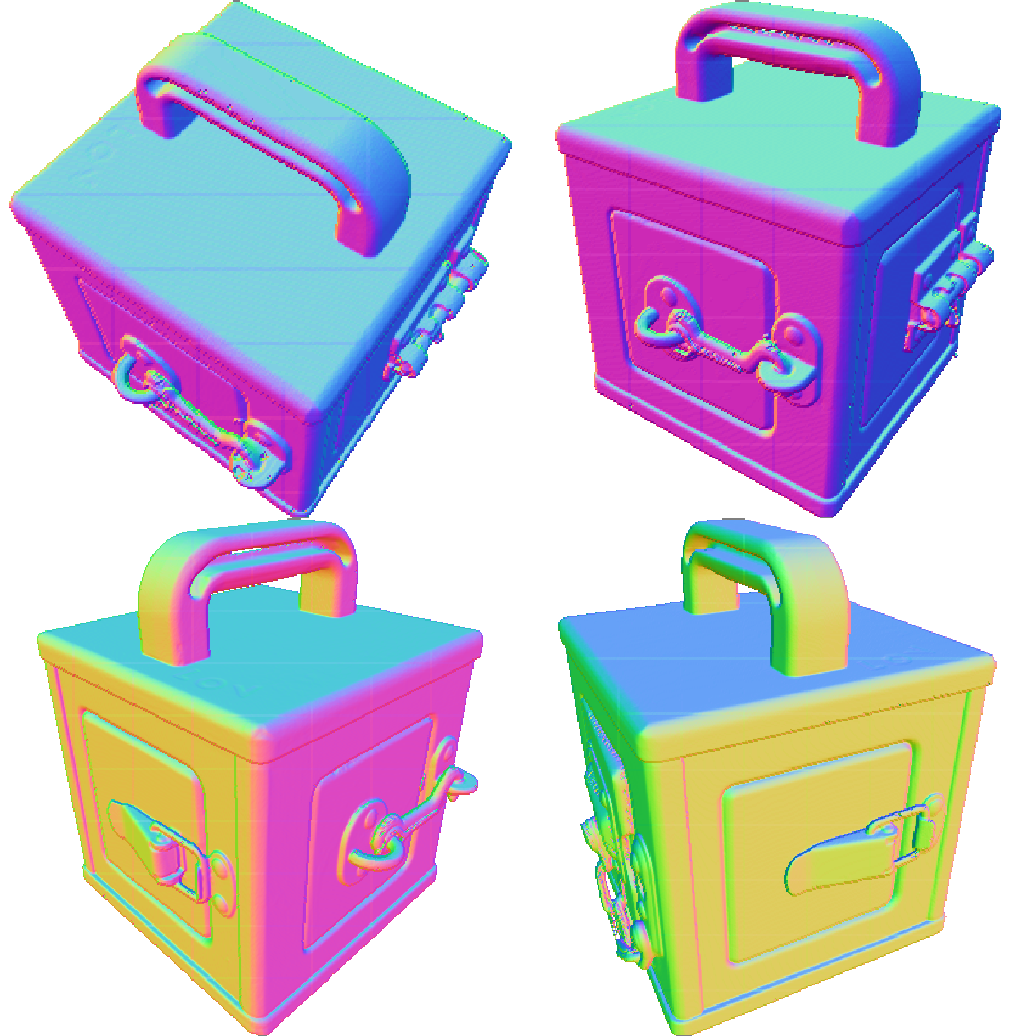} &
        \includegraphics[width=0.19\textwidth]{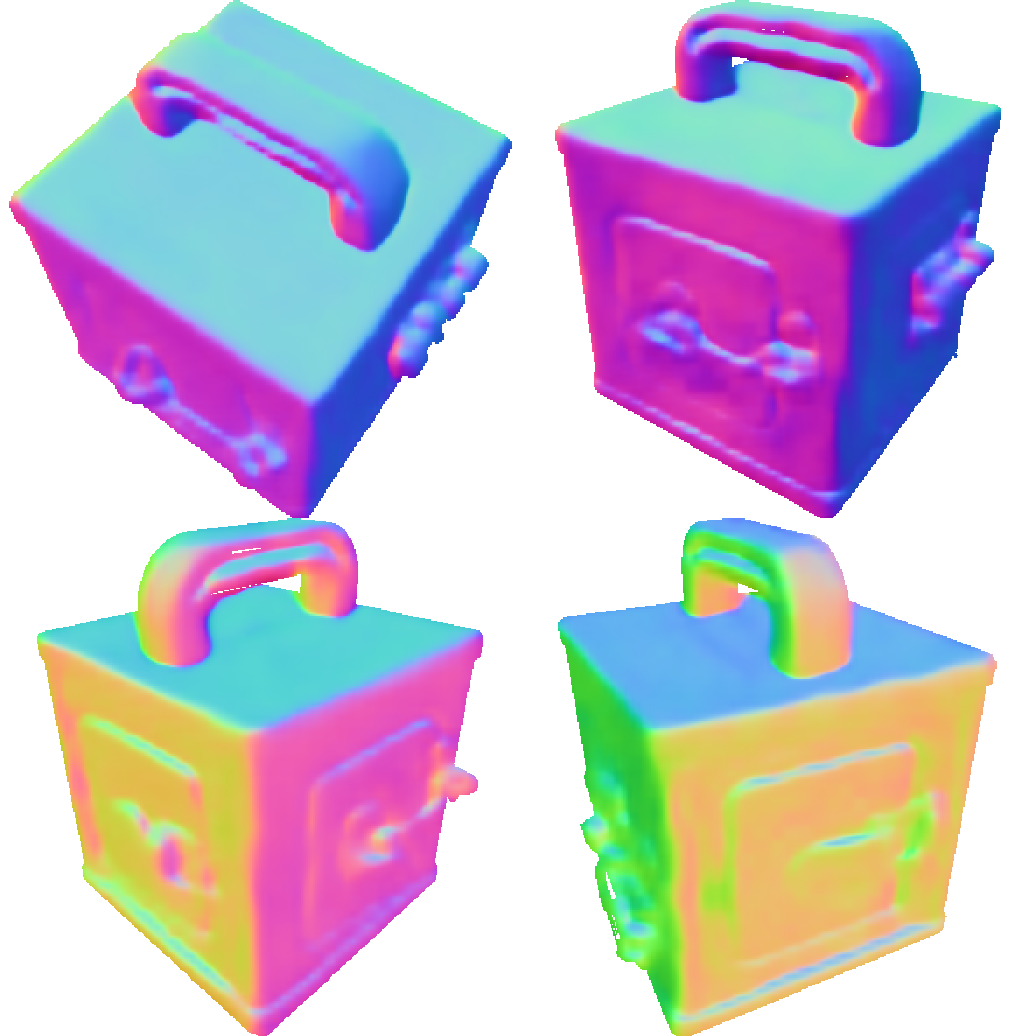} &
        \includegraphics[width=0.19\textwidth]{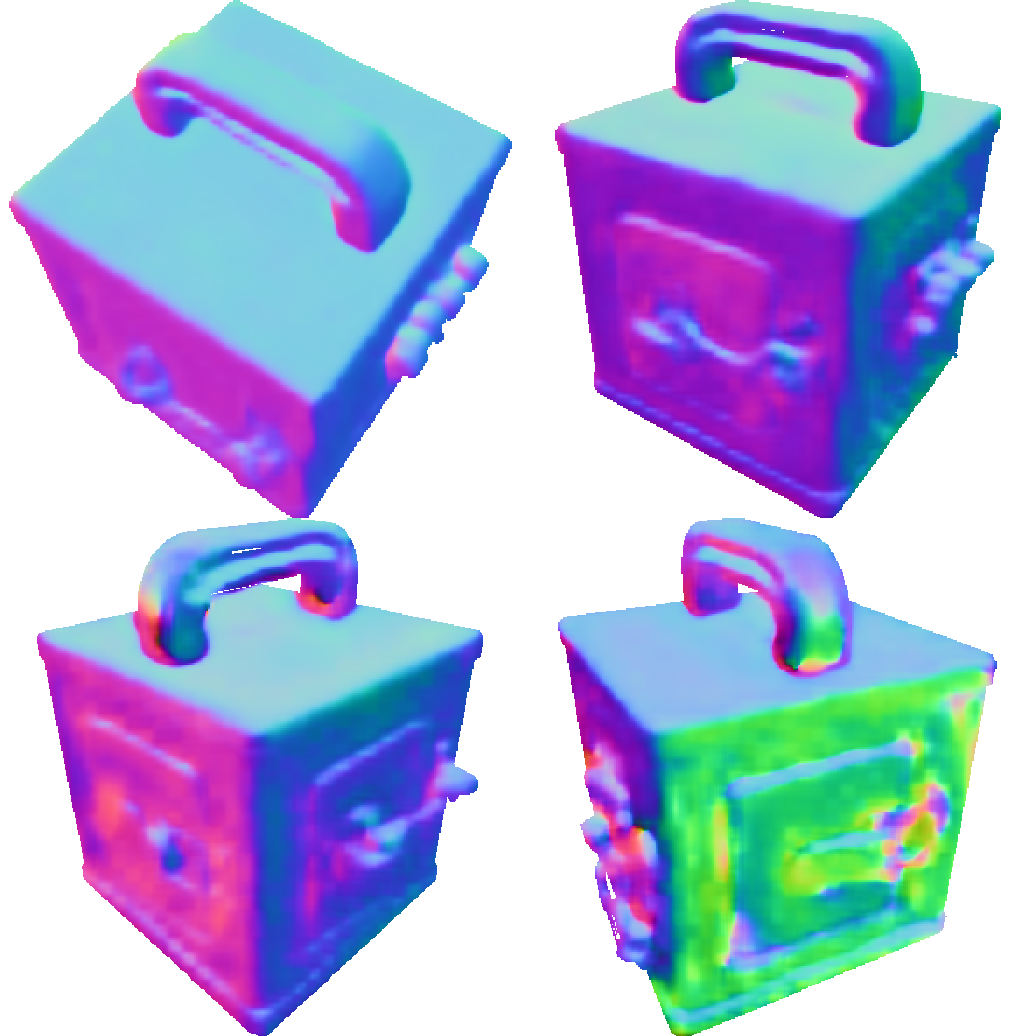} &
        \includegraphics[width=0.19\textwidth]{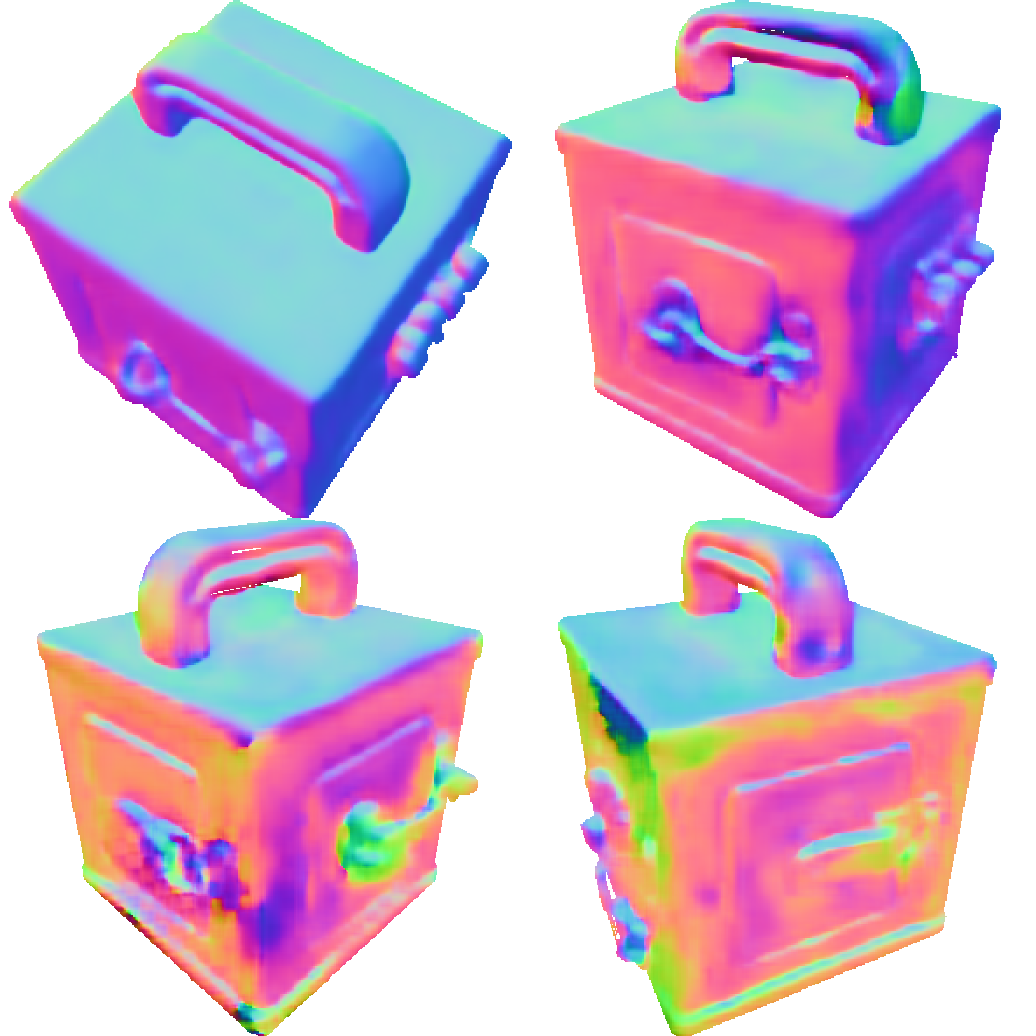} \\

        \includegraphics[width=0.19\textwidth]{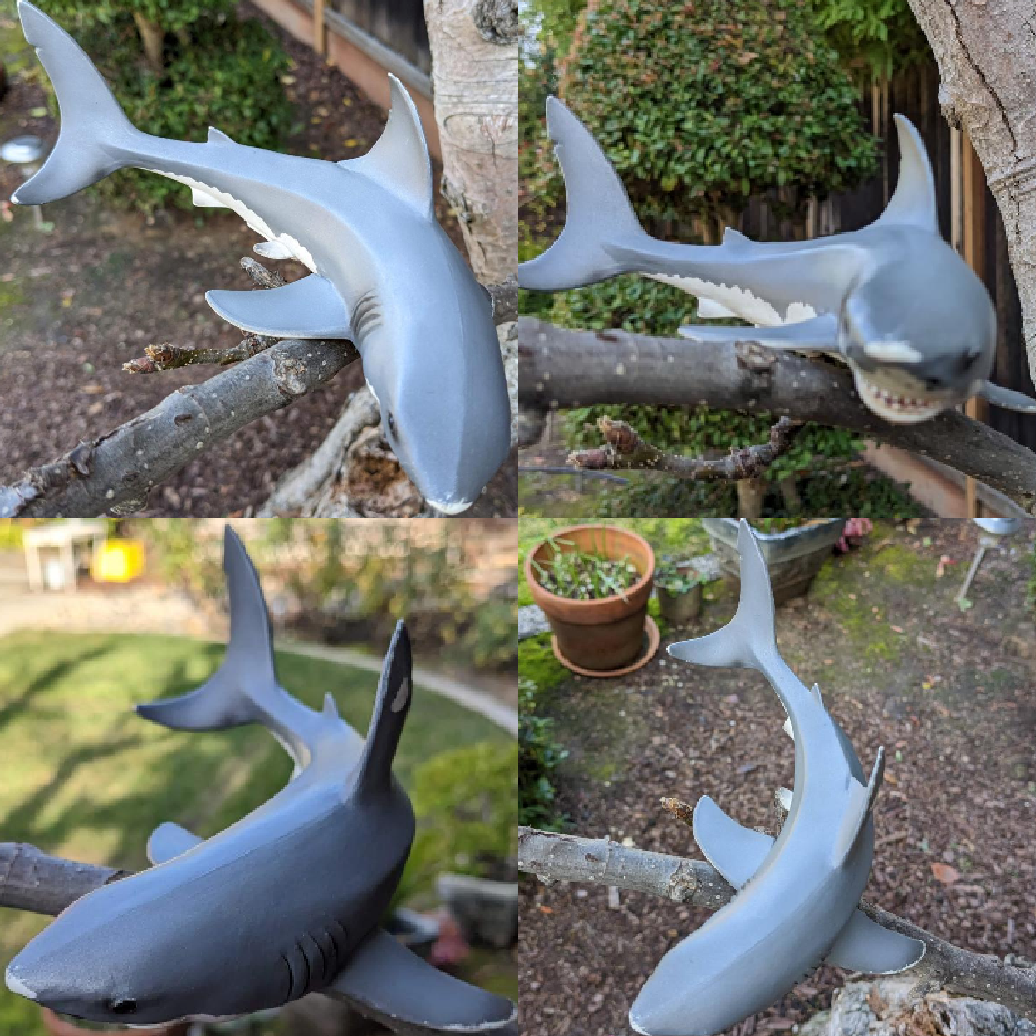} &
        \includegraphics[width=0.19\textwidth]{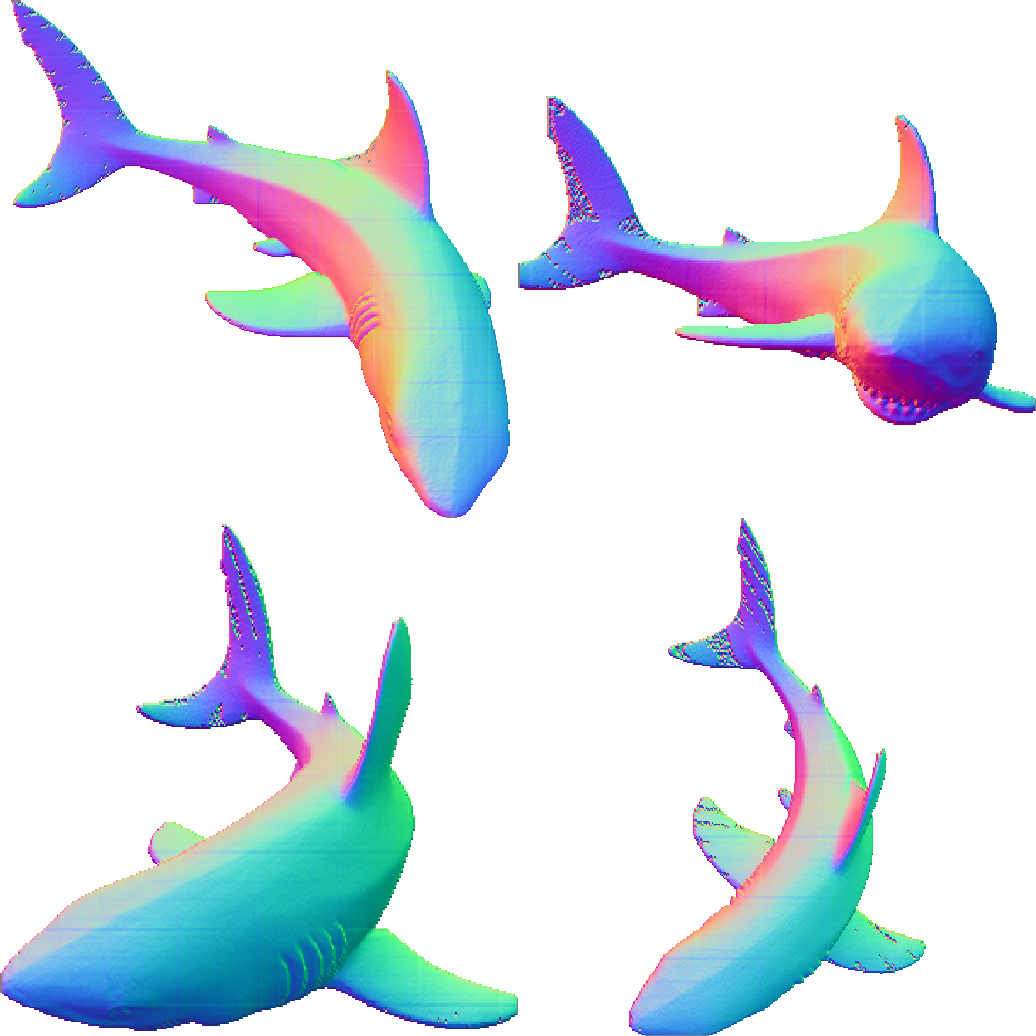} &
        \includegraphics[width=0.19\textwidth]{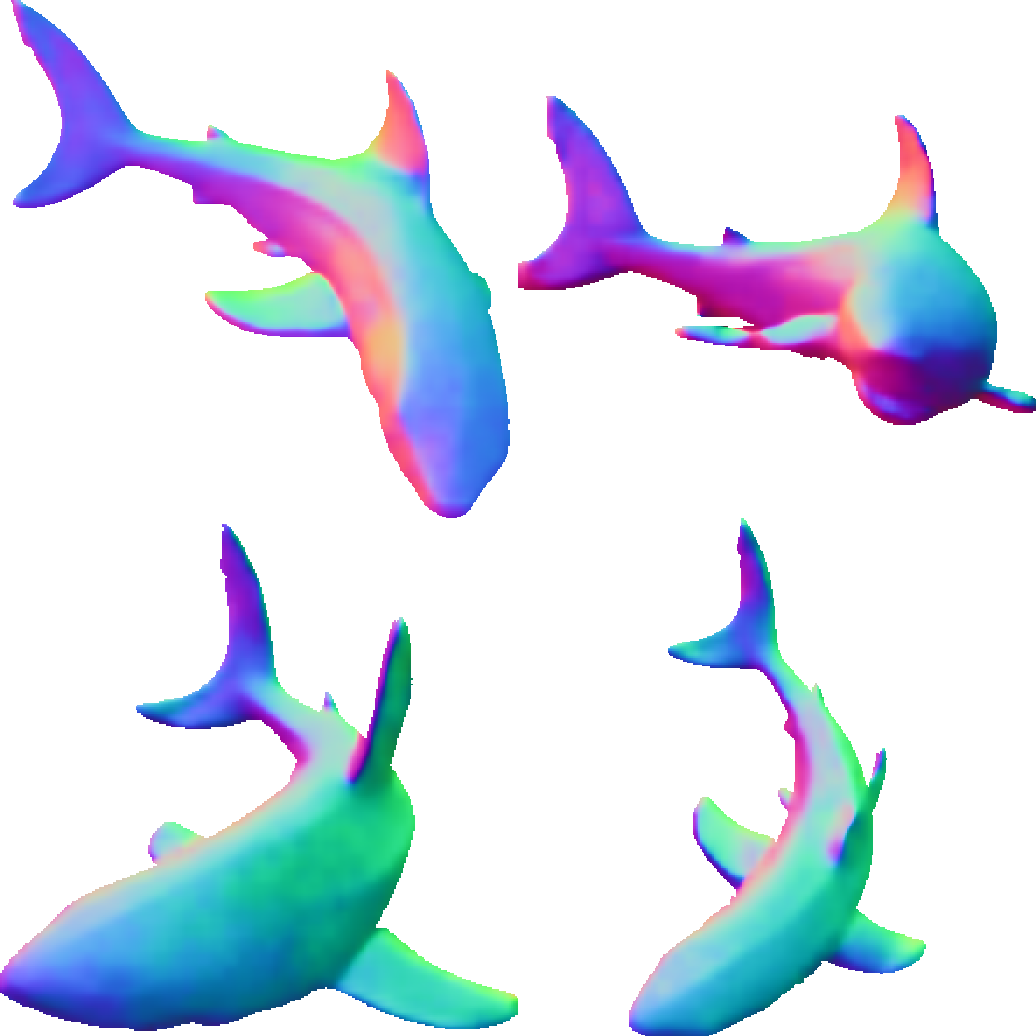} &
        \includegraphics[width=0.19\textwidth]{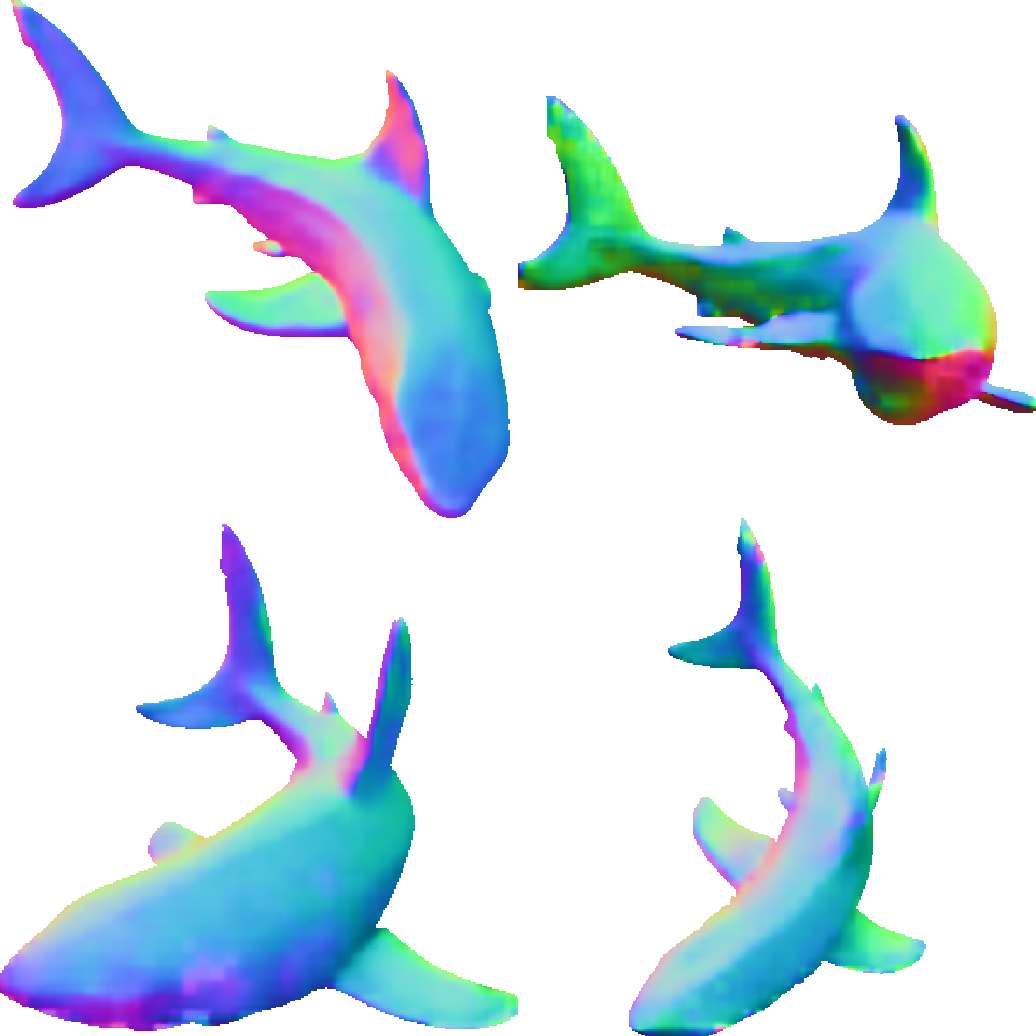} &
        \includegraphics[width=0.19\textwidth]{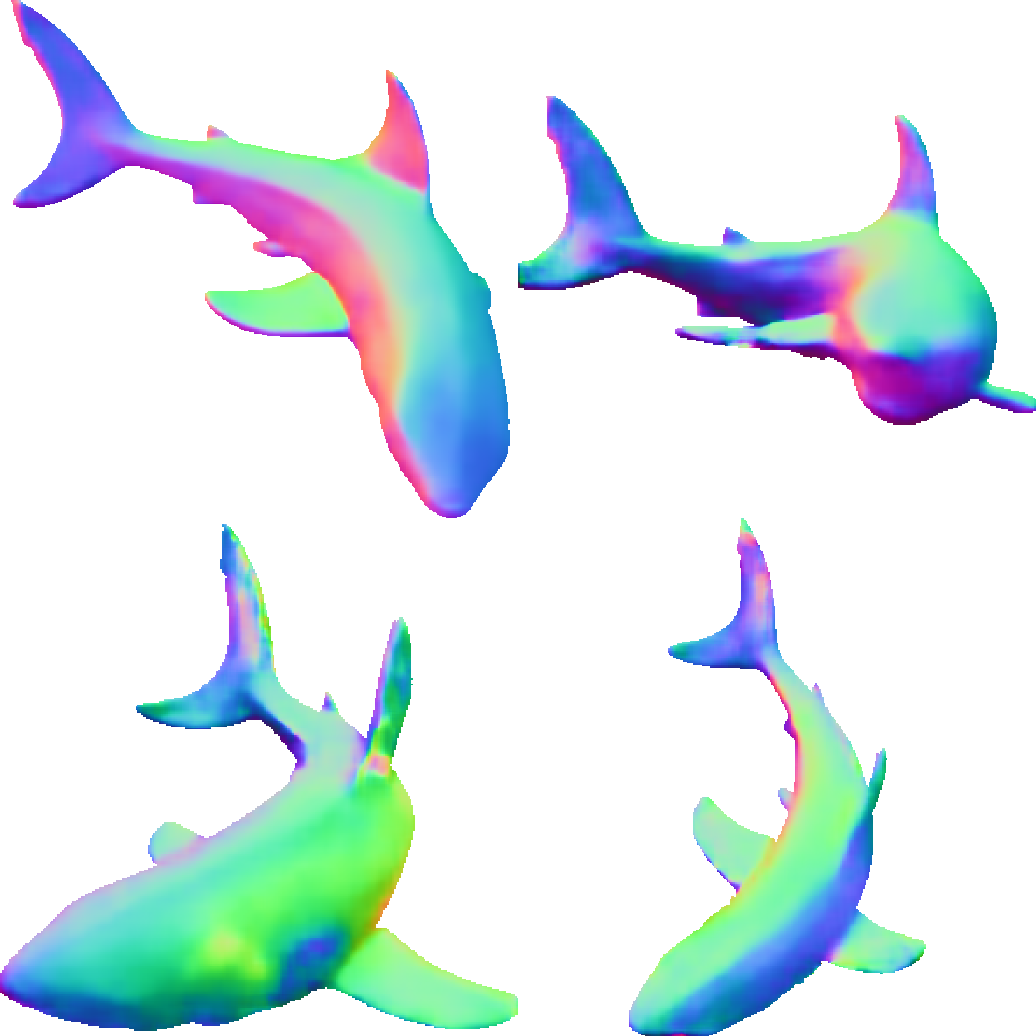} \\

    \end{tabular}
    \caption{\textbf{More Qualitative Results for 3D-Aligned Surface Normal Estimation.} We visualize the predicted surface normals from multiple viewpoints. The consistency in color and orientation across different views demonstrates that our multi-view features encode a coherent 3D representation. Note how surfaces facing the same world direction retain the same color regardless of the camera angle. 
    }
    \label{fig:snorm}

\end{figure*}

We also differentiate our problem setting from 3D foundation models.
Unlike reconstruction pipelines (VGGT, DUSt3R, $\pi^3$) that regress explicit geometry or specialized tools (PANSt3R), our contribution is a \textbf{feature-space adaptation} that lifts 2D foundation models into multi-view-consistent backbones while \textbf{preserving native semantics}. Table~\ref{tab:comparison} validates this: while reconstruction models prioritize geometry, they destroy the semantic feature space. Our method, in contrast, preserves backbone semantics (demonstrated by zero-shot classification) and endows them with multi-view consistency (demonstrated by localization metrics). Crucially, \textbf{we do not aim to surpass specialized solvers on specific tasks}, but to provide versatile backbones. 
Correspondence serve as a metric to measure multi-view consistency, not as a target application. Thus, the appropriate comparison is to general lifting frameworks (e.g., Fit3D) rather than narrow task specialists.
\begin{table}[t]
    \centering
    \captionof{table}{\textbf{Point Cloud Estimation.} Our method yields the most accurate PC reconstruction on NAVI dataset.}
    \label{tab:results_nocrop}
    \footnotesize
    \setlength{\tabcolsep}{3pt}
    \begin{tabular}{l c c c}
    \toprule
     & RMSE $\downarrow$  & Chamfer ($10^{-2}$) $\downarrow$ \\
    \midrule
    Base  & 0.0708  & 3.76 \\
    SND   & 0.0720  &  3.96 \\
    Fit3D & 0.0814  & 4.39 \\
    MEF   & 0.0814  &  4.93 \\
    \midrule
    \textbf{Ours} &  \textbf{0.0644} & \textbf{3.29} \\
    \bottomrule
    \end{tabular}
\end{table}
\begin{table}
\centering
\scriptsize
\captionof{table}{
\textbf{Reconstruction vs. Lifting.} We fundamentally distinguish our approach as a \textit{lifting} method - applicable to DINO, SAM, and CLIP - from \textit{reconstruction} specialists (VGGT, DUSt3R). While reconstruction models prioritize geometry, they effectively destroy the foundation model itself, evidenced by the catastrophic failure on ImageNet KNN classification (\textcolor{red}{\textbf{marked in red}}). Our framework tackles a different goal: we introduce multi-view consistency while strictly preserving the native semantic capabilities of the original backbone.
"VGGT - encoder" is DINOv2 backbone + Alternating-Attention encoder, "Full Pipeline" adds a DPT decoder and CoTracker2 model.
}
\begin{tabular}{l|c|cc|cc|cc}
\hline
 \multirow{2}{*}{\textbf{Method}} 
 & \multirow{2}{*}{\textbf{\#Params}}
 & \multicolumn{2}{c}{\textbf{KNN Zero-Shot}} 
 & \multicolumn{2}{c}{\textbf{ScanNet++}} 
 & \multicolumn{2}{c}{\textbf{Generalization}} \\
 & 
 & \textbf{Top-1} & \textbf{Top-5} 
 & \textbf{Loc. Err} $\downarrow$ & \textbf{Base Sim.} $\uparrow$ 
 & \textbf{Loc. Err} $\downarrow$ & \textbf{Base Sim.} $\uparrow$ \\
\hline
Base DINOv2 & 22M
 & 62.50\% & 84.37\%  
 & 0.1029 & -  
 & 0.1404 & -  \\
 \hdashline
VGGT - encoder & 909M
 & \colorbox{red!50}{8.39\%} & \colorbox{red!50}{18.08\%}  
 & 0.1140 & \colorbox{red!50}{0.1636}  
 & 0.0899 & \colorbox{red!50}{0.2322}   \\
VGGT - Full & 1.25B
 & N/A & N/A  
 & 0.0327 & N/A
 & \textbf{0.0259} & N/A  \\
\textbf{Ours} & \textbf{28M}
 & \textbf{58.02}\% & \textbf{81.21\%}  
 & \textbf{0.0247} & \textbf{0.9376}
 & 0.0782 & \textbf{0.8003}  \\
 \hline
\end{tabular}
\label{tab:comparison}
\end{table}%

\section{Implementation Details}
All experiments ran on a single NVIDIA RTX 5090 (32\,GB). We trained on the ScanNet++ training split for 24 epochs (SAM and DINOv3 for 12 epochs) with batch size 4 and set size 4, which means total of 16 images per batch. Per scene, each epoch sampled 1{,}000 sets of 4 images (512$\times$512) with ground-truth correspondences, and at each iteration, the correspondence loss was computed over 128 correspondences sampled across the set. Optimization used AdamW (lr = 1e$^{-4}$) with a linear learning-rate scheduler. Fine-tuning employed LoRA with rank $r{=}32$ and $\alpha{=}32$. Backbones: DINOv2 ViT-S/14 with register tokens (\texttt{vits14\_reg}) following Darcet~\etal, SAM ViT-B, and DINOv3 and CLIP with ViT-B/16. For the multi-view surface-normal probe, we froze all backbone weights and trained a self-attention-based multi-view head to predict per-pixel normals aligned to a single 3D reference frame (camera 0 as origin). The probe produced four channels $(x,y,z,\sigma)$ and was trained with an uncertainty-weighted loss following Banani~\etal, and the $(x,y,z)$ outputs were $\ell_2$-normalized to unit vectors. Probe training used AdamW for 50 epochs (lr = 5e$^{-5}$) with a LambdaLR schedule.

For all baseline methods, we used the officially released code and pre-trained weights.


\end{document}